%% file: main.tex
\documentclass[twoside]{article}

\usepackage[accepted]{aistats2026}
\input{001_preamble}
\input{002_macro}

\begin{document}



\twocolumn[

\aistatstitle{Free Random Projection for In-Context Reinforcement Learning}

\aistatsauthor{ Tomohiro Hayase \And Beno\^{\i}t Collins \And  Nakamasa Inoue }

\aistatsaddress{ AIST and CoeFont Co., Ltd. \And  Kyoto University \And Institute of Science Tokyo } ]

\input{009_abst_intro}

\input{010_body}

\input{070_discussion}

\input{080_acknowledgement}

%\bibliographystyle{abbrvnat}
%\bibliographystyle{unsrtnat}
%\bibliographystyle{plainnat}
%\bibliographystyle{natbib}
%\setcitestyle{authoryear,open={((},close={))}} %Citation-related commands

\bibliographystyle{hunsrtnatarxiv} %% for unsort-natbib with arXiv
\bibliography{reference}

\section*{Checklist}

% %%% BEGIN INSTRUCTIONS %%%
%The checklist follows the references. For each question, choose your answer from the three possible options: Yes, No, Not Applicable.  You are encouraged to include a justification to your answer, either by referencing the appropriate section of your paper or providing a brief inline description (1-2 sentences). 
%Please do not modify the questions.  Note that the Checklist section does not count towards the page limit. Not including the checklist in the first submission won't result in desk rejection, although in such case we will ask you to upload it during the author response period and include it in camera ready (if accepted).

%\textbf{In your paper, please delete this instructions block and only keep the Checklist section heading above along with the questions/answers below.}
% %%% END INSTRUCTIONS %%%

\begin{enumerate}

  \item For all models and algorithms presented, check if you include:
  \begin{enumerate}
    \item A clear description of the mathematical setting, assumptions, algorithm, and/or model. [Yes]
    \item An analysis of the properties and complexity (time, space, sample size) of any algorithm. [Yes]
    \item (Optional) Anonymized source code, with specification of all dependencies, including external libraries. [Yes]
  \end{enumerate}

  \item For any theoretical claim, check if you include:
  \begin{enumerate}
    \item Statements of the full set of assumptions of all theoretical results. [Yes]
    \item Complete proofs of all theoretical results. [Yes]
    \item Clear explanations of any assumptions. [Yes]     
  \end{enumerate}

  \item For all figures and tables that present empirical results, check if you include:
  \begin{enumerate}
    \item The code, data, and instructions needed to reproduce the main experimental results (either in the supplemental material or as a URL). [Yes]
    \item All the training details (e.g., data splits, hyperparameters, how they were chosen). [Yes]
    \item A clear definition of the specific measure or statistics and error bars (e.g., with respect to the random seed after running experiments multiple times). [Yes]
    \item A description of the computing infrastructure used. (e.g., type of GPUs, internal cluster, or cloud provider). [Yes]
  \end{enumerate}

  \item If you are using existing assets (e.g., code, data, models) or curating/releasing new assets, check if you include:
  \begin{enumerate}
    \item Citations of the creator If your work uses existing assets. [Yes]
    \item The license information of the assets, if applicable. [Not Applicable]
    \item New assets either in the supplemental material or as a URL, if applicable. [Not Applicable]
    \item Information about consent from data providers/curators. [Not Applicable]
    \item Discussion of sensible content if applicable, e.g., personally identifiable information or offensive content. [Not Applicable]
  \end{enumerate}

  \item If you used crowdsourcing or conducted research with human subjects, check if you include:
  \begin{enumerate}
    \item The full text of instructions given to participants and screenshots. [Not Applicable]
    \item Descriptions of potential participant risks, with links to Institutional Review Board (IRB) approvals if applicable. [Not Applicable]
    \item The estimated hourly wage paid to participants and the total amount spent on participant compensation. [Not Applicable]
  \end{enumerate}

\end{enumerate}

\clearpage
\appendix
\thispagestyle{empty}

% Supplementary material: To improve readability, you must use a single-column format for the supplementary material.
\onecolumn
\aistatstitle{Supplementary Materials}
\input{090_appendix}

\end{document}

%% file: 001_preamble.tex
\usepackage{natbib}

\usepackage[utf8]{inputenc} % allow utf-8 input
\usepackage[T1]{fontenc}    % use 8-bit T1 fonts
\usepackage{hyperref}       % hyperlinks
\usepackage{url}            % simple URL typesetting
\usepackage{booktabs}       % professional-quality tables
\usepackage{amsfonts}       % blackboard math symbols
\usepackage{nicefrac}       % compact symbols for 1/2, etc.
\usepackage{microtype}      % microtypography
\usepackage{xcolor}         % colors

\usepackage{amsthm}
\usepackage{amsmath}
\usepackage{amssymb} % for curvearrowright
\usepackage{mathrsfs} % for analytic S-transform
\usepackage[]{graphicx} % for pdflatex

\usepackage{zref-clever}

\let\cref\zcref
\let\Cref\zcref

\zcsetup{
  cap    = true,   % capitalize type names
  abbrev = false   % no abbreviations
}

\zcRefTypeSetup{equation}{
  Name-sg = {}, name-sg = {},
  Name-pl = {}, name-pl = {},
  refbounds-first = {(, ), (, )},
  refbounds-last  = {(, ), (, )},
}

\AddToHook{env/thm/begin}{\zcsetup{countertype={thm=theorem}}}
\AddToHook{env/prop/begin}{\zcsetup{countertype={thm=proposition}}}
\AddToHook{env/lemma/begin}{\zcsetup{countertype={thm=lemma}}}
\AddToHook{env/cor/begin}{\zcsetup{countertype={thm=corollary}}}
\AddToHook{env/question/begin}{\zcsetup{countertype={thm=remark}}}
\AddToHook{env/hyp/begin}{\zcsetup{countertype={thm=hypothesis}}}
\AddToHook{env/definition/begin}{\zcsetup{countertype={thm=definition}}}
\AddToHook{env/remark/begin}{\zcsetup{countertype={thm=remark}}}
\AddToHook{env/example/begin}{\zcsetup{countertype={thm=example}}}
\AddToHook{env/concl/begin}{\zcsetup{countertype={thm=result}}}
\AddToHook{env/note/begin}{\zcsetup{countertype={thm=note}}}
\AddToHook{env/conj/begin}{\zcsetup{countertype={thm=proposition}}}
\AddToHook{env/assumption/begin}{\zcsetup{countertype={thm=theorem}}}

\renewcommand{\ref}{\zcref}
\usepackage{comment}

\usepackage{algorithm}
\usepackage{algpseudocode}

\theoremstyle{plain}
\newtheorem{thm}{Theorem}[section]

\newtheorem{lemma}[thm]{Lemma}

\theoremstyle{definition}
\newtheorem{definition}[thm]{Definition}

\numberwithin{equation}{section}

\renewcommand{\paragraph}[1]{\noindent{\bf #1.}}
\usepackage{wrapfig} % wraptable
\usepackage{caption} % for \captionsetup
\usepackage{multirow} % for multirow in table

%% file: 002_macro.tex
\newcommand{\N}{\mathbb{N}}
\newcommand{\R}{\mathbb{R}}
\newcommand{\C}{\mathbb{C}}
\newcommand{\E}{\mathbb{E}}

\DeclareMathOperator{\Tr}{\mathrm{Tr}}

\DeclareMathOperator{\supp}{\mathrm{supp}}

\newcommand{\tp}{\mathsf{T}}
\usepackage{tikz}

\newcommand{\F}{\mathbb{F}}

\newcommand{\rep}{\lambda} % representation of F_n
\newcommand{\scaling}{\sigma_w} % scaling factor of FRP
\newcommand{\eoe}{1^\mathrm{done}} % end of environment
\newcommand{\length}{\ell}
\newcommand{\genset}{\Lambda}
\newcommand{\ngen}{n}  % the number of elements of base
\newcommand{\nwords}{n_w}
\newcommand{\nenv}{{n_e}} % the number of environments
\newcommand{\orthg}[1]{{\mathrm{O}({#1})}}
\newcommand{\permg}[1]{{\mathrm{S}({#1})}}
\newcommand{\words}{\genset_\ngen^\length}
\newcommand{\states}{\mathcal{S}}
\newcommand{\Unif}{\mathrm{Unif}}
\newcommand{\env}{\mathcal{E}}
\newcommand{\obs}{\xi}% observation vector
\newcommand{\sphere}[1]{S^{#1-1}} % d-1 dimensional sphere

\newcommand{\drp}{d}

\newcommand{\din}{d_\mathrm{in}} % dimension of input vector of embedding network
\newcommand{\dobs}{d_\mathrm{obs}} % dimension of environment
\newcommand{\worddist}{\rho_\F}

\newcommand{\EmbNN}{f^\mathrm{emb}_{\theta^e}}%{G^\mathrm{E}}
\newcommand{\RUnit}{F_{\theta^u}}
\newcommand{\dhidden}{d_h}
\newcommand{\RNN}{G_{\theta^r}}
\newcommand{\PolicyNet}{f^\mathrm{policy}_{\theta^p}}
\newcommand{\ValueNet}{f^\mathrm{value}_{\theta^v}}
\newcommand{\act}{\rho}

\newcommand{\effdim}{d_\mathrm{eff}}
\newcommand{\kermat}{K}

\usepackage{color}
\usepackage[normalem]{ulem}

%% file: 009_abst_intro.tex
\begin{abstract}
Hierarchical inductive biases are hypothesized to promote generalizable policies in reinforcement learning, as demonstrated by explicit hyperbolic latent representations and architectures. Therefore, a more flexible approach is to have these biases emerge naturally from the algorithm. We introduce Free Random Projection, an input mapping grounded in free probability theory that constructs random orthogonal matrices where hierarchical structure arises inherently. The free random projection integrates seamlessly into existing in-context reinforcement learning frameworks by encoding hierarchical organization within the input space without requiring explicit architectural modifications. Empirical results on multi-environment benchmarks show that free random projection consistently outperforms the standard random projection, leading to improvements in generalization. Furthermore, analyses within linearly solvable Markov decision processes and investigations of the spectrum of kernel random matrices reveal the theoretical underpinnings of free random projection's enhanced performance, highlighting its capacity for effective adaptation in hierarchically structured state spaces.
\end{abstract}

\section{Introduction}\label{sec:intro}
Despite major progress in deep reinforcement learning (RL)~\citep{sutton2018reinforcement}, generalization across multi-environment or multi-task settings remains a bottleneck.
\emph{In-context reinforcement learning} (ICRL)~\citep{lin2023transformers,Laskin2023context,Lee2023supervised,lu2023structured,sinii2024context} addresses this by enabling agents to adapt within their context window at decision time without updating parameters. By processing sequences of states, actions, and rewards, agents can infer task specifications on the fly, allowing rapid adaptation with minimal data. 
In \citep{kirsch2022general,lu2023structured,sinii2024context}, random projections map input vectors into a unified dimension to treat them with a single model.
However, discovering the effective processing that accommodates diverse state/action specifications while preserving essential structure remains challenging.

Many tasks in RL exhibit a tree-like or hierarchical structure, with exponential branching in possible futures. Recent research indicates that hyperbolic geometry can effectively capture these hierarchical relationships in RL~\citep{cetin2023hyperbolic} and graph-based tasks~\citep{nickel2017Poincare, chami2019hyperbolic}. Yet, most approaches that leverage hyperbolic geometry rely on explicit latent representation spaces or model architectures. Therefore, it would be more flexible to have the learning algorithm itself induce hierarchical inductive biases rather than rely solely on dedicated model designs.

To address these challenges, we introduce \emph{Free Random Projection} (FRP), a sampling rule that replaces the standard i.i.d.\ orthogonal projection with a structured distribution induced by words in a free group. Here, the relevant hierarchy is not a literal tree over the sequence of sampled environments. Rather, the words form a prefix tree, and two words that share a long prefix yield projection matrices whose defining products share the corresponding prefix of the base orthogonal matrices. This gives the projection family a shared-prefix dependence structure that is absent in standard random projection, while leaving the policy architecture unchanged. Our theoretical analysis studies this representation-side effect through the induced kernel and its effective dimension, whereas our RL experiments test whether the resulting inductive bias can improve cross-environment generalization in the ICRL setting.

Our contributions are primarily as follows.
\begin{enumerate}
    \item Kernel analysis (\cref{ssec:average-kernel-analysis}) shows that FRP induces a different joint distribution from standard random projection and lowers the effective dimension of the induced kernel (\cref{thm:eff-dim}), providing a representation-level mechanism that may contribute to improved cross-environment generalization.
    \item In experiments on multi-environment RL tasks (\cref{sec:pomdps}), FRP consistently outperforms standard random projection approaches, demonstrating its suitability for in-context adaptation. 
    \item We further analyze FRP in linearly solvable MDPs (LSMDPs), revealing when FRP better exploits hierarchical state-space structure in policy learning (\cref{ssec:lsmdp}).
\end{enumerate}
 Overall, these findings indicate that \textbf{ incorporation of hierarchical properties within a random projection framework} can yield a beneficial implicit bias for ICRL.

\begin{figure*}[t]
    \centering    
    \includegraphics[width=0.28\linewidth]{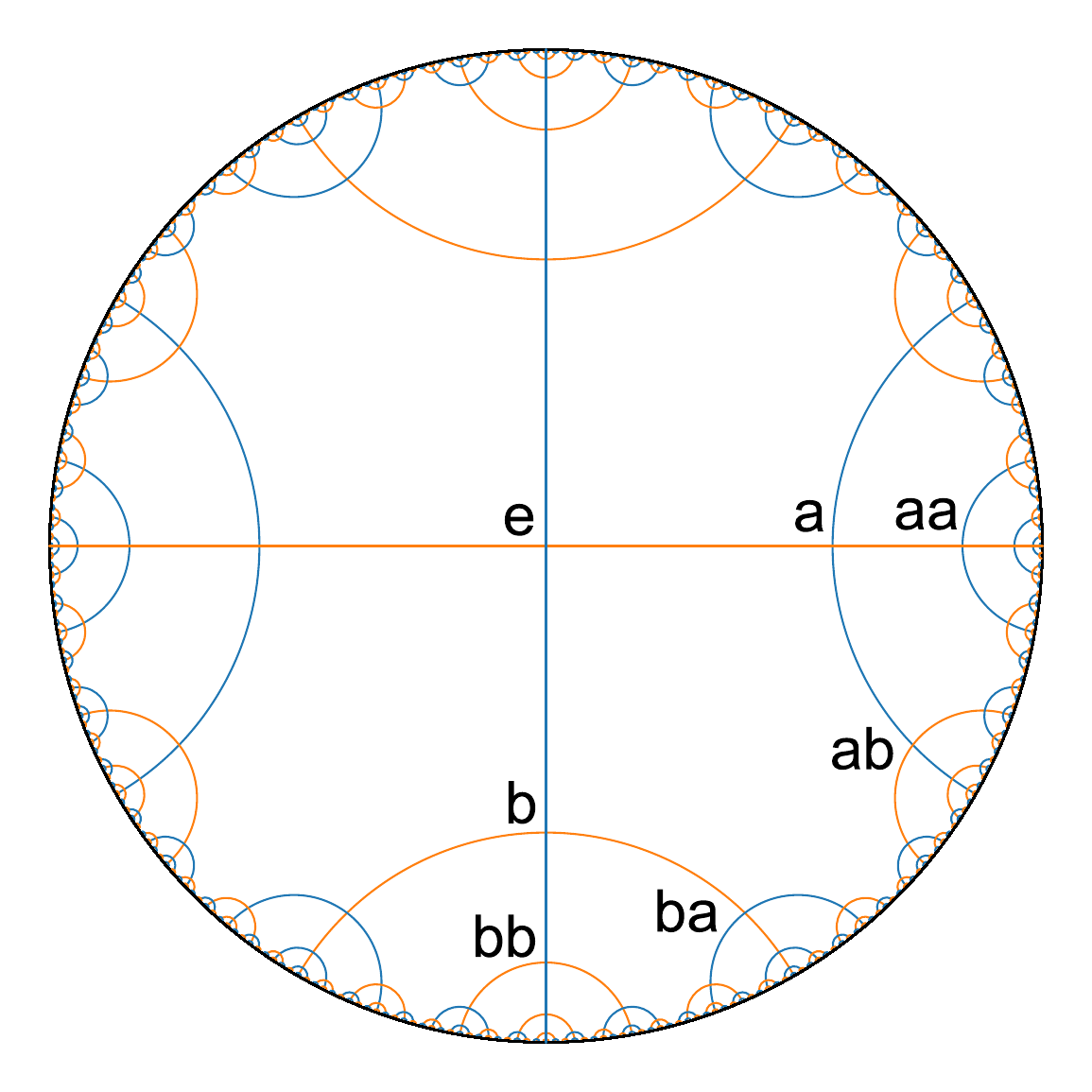}
    \includegraphics[width=0.7\linewidth]{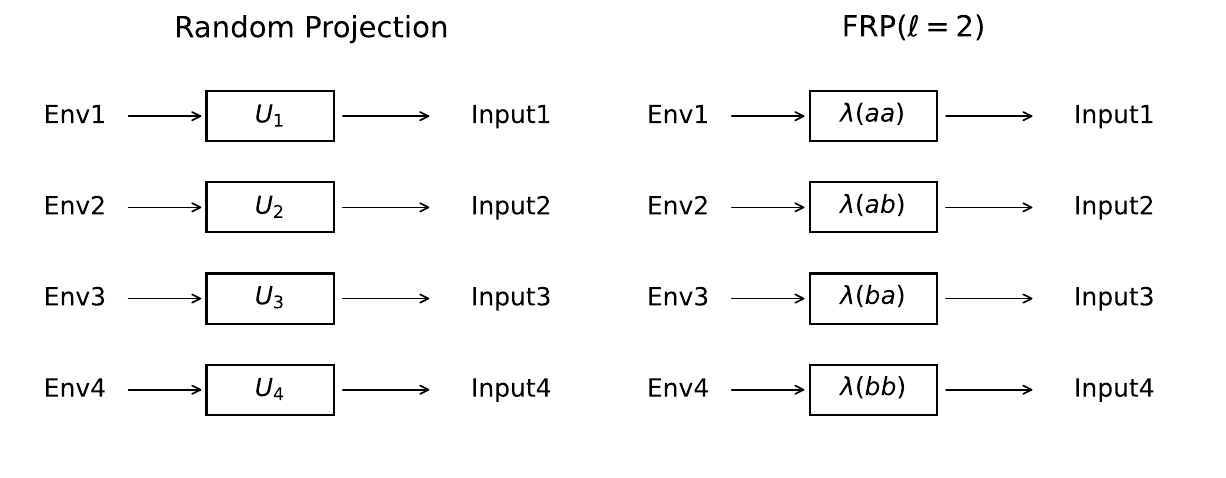}
    \caption{Overview of FRP. (Left) The Cayley graph of the free group $\F_2$, shown in the Poincar\'e disk as a visualization of the prefix-tree structure of reduced words. (Center) Standard random projection, which applies i.i.d.\,mappings to each environment. (Right) FRP using words in $\F_2$; words with longer shared prefixes yield matrix products with longer common prefixes.}
    \label{fig:overview}
\end{figure*}

\subsection{Related Work}
\paragraph{Hierarchical or Hyperbolic Bias}
Hyperbolic or tree-based embeddings can capture hierarchical relationships~\citep{nickel2017Poincare,nickel2018learning,chami2019hyperbolic}. In RL, \citet{cetin2023hyperbolic} introduced a hyperbolic latent representation for exponentially branching transitions, though most of the existing work in hyperbolic RL relies on specialized architectures or loss functions. In contrast, FRP's hierarchical bias arises from \textbf{altering the random projection distribution}, enabling integration into standard pipelines without architectural modification. This approach also aligns with recent findings that \emph{implicit biases} in training can improve generalization without explicit model architecture design~\citep{soudry2018implicit, arora2019implicit,lyu2020gradient,nakkiran2020deep,Shah2020Pitfalls,martin2021implicit,Nacson2022Implicit}.

\paragraph{Random Projections and Matrices in ML} Random projections effectively support dimensionality reduction and data augmentation by conserving fundamental geometric features of datasets while requiring minimal parameter adjustments. Beyond traditional use in Johnson-Lindenstrauss embeddings, insights from random matrix theory (RMT) have expanded understanding in areas such as neural network initialization, signal propagation, and kernel analysis \citep{saxe2014exact,Pennington2017nonlinear,jacot2018neural,fan2020spectra,martin2021implicit,wang2024deformed}.
Free probability theory (FPT)~\citep{voiculescu1992free, mingo2017free} presents an algebraic perspective on random matrices, offering strong methods for spectral or kernel analysis \citep{pennington2017resurrecting,pennington2018emergence,xiao2018dynamical,hanin2020products,chhaibi2022free,hayase2021spectrum,naderi2024mind}. The use of RMT \citep{ilbert2024analysing} or FPT \citep{dobriban2019asymptotics, patil2024asymptotically, hayase2020cauchy} in machine learning algorithms is notable. In RL, random matrices facilitate applications such as exploration bonuses \citep{Burda2019Exploration} and the formation of random kernel matrices \citep{chen2023self, lee2023spqr}. Furthermore, random projections are increasingly applied in meta-RL and ICRL, particularly to standardize observation or action spaces \citep{kirsch2022general,lu2023structured,sinii2024context}. These varied uses illustrate how random matrices contribute to improved stability, efficiency, and task generalization. 
Drawing from these foundations, FRP uses free-group words to organize a projection family with shared-prefix dependence; to our knowledge, this representation-side use of free probability has not been explored in machine learning theory.

%\paragraph{ICRL}
%ICRL \citep{kirsch2022general,lu2023structured,sinii2024context} allows agents to adapt to new tasks using their context window (trajectories of states, actions, and rewards), rather than requiring parameter updates. This approach has shown success with RNN or transformer-based models that learn entire algorithms within their weights. Our work extends these ideas by showing that \textbf{altering the random projection distribution} alone—when augmented with hierarchical and hyperbolic structure—can also facilitate in-context adaptation. This highlights that emergent biases, rather than explicit architectural design alone, can be crucial for rapid adaptation and generalization. 

\section{Preliminaries}

\subsection{Reinforcement Learning Setup}
As a standard setting for reinforcement learning, we consider a standard Markov Decision Process (MDP)~\citep{sutton2018reinforcement} defined by the tuple $(\states, \mathcal{A}, P, R, \gamma)$. Here, $\states$ is the set of states; $\mathcal{A}$ is the set of actions; $P(s' \mid s, a)$ is the state transition probability from $s$ to $s'$ under action $a$; $R(s,a)$ is the reward function; and $\gamma \in [0,1)$ is the discount factor. The agent's objective is to learn a policy $\pi(a \mid s)$ that maximizes the expected discounted return $J(\pi)=\mathbb{E}_\pi[\sum_{t=0}^{\infty}\gamma^t R(s_t, a_t)]$.

In partially observable (POMDP) settings, the agent only has access to an observation $\obs_t$ drawn from an observation function $P_o(\cdot \mid s_t)$ rather than the state $s_t$. Consequently, the agent's policy often depends on the entire history of observations and actions $(\obs_0,a_0,\obs_1,a_1,\dots)$. 

\subsection{Meta-Reinforcement Learning and ICRL}
\label{sec:meta-incontext}
Meta-reinforcement learning (meta-RL) targets fast adaptation across a distribution of tasks \citep{beck2023survey}. In many conventional meta-RL methods, %tasks share a common observation and action space, and 
the agent's objective is to quickly learn a near-optimal policy for each new task using knowledge gained from prior tasks.
\emph{In-context reinforcement learning} (ICRL) extends meta-RL by enabling adaptation \emph{within} the agent's context window at decision time, eliminating the need for parameter updates \citep{Laskin2023context,Lee2023supervised}. By processing consecutive trajectories of observations, actions, and rewards, the agent infers the underlying task-specific properties on the fly. %This capability is particularly valuable in partially observable or multi-environment scenarios, where effective latent representation and rapid adaptation are crucial.

\subsection{Random Projection in Meta-RL}
A key challenge in meta-RL is dealing with heterogeneous observation or action spaces across multiple tasks or environments. An approach is to embed these variable-dimensional inputs into a unified dimensional space using a random projection~\citep{kirsch2022general,lu2023structured, sinii2024context}. Formally, an input observation vector $\obs \in \mathbb{R}^{\dobs}$ is mapped to $\obs' \in \mathbb{R}^{\din}$ by a uniformly distributed isometry $M_o$, decomposed as $\obs' = M_o\,\obs = \scaling T_2\,U\,T_1\,\obs$, where $\scaling>0$ is a scaling factor, $T_1$ and $T_2$ are fixed rectangular identity matrices used to handle dimension mismatch; and $U$ is an orthogonal matrix drawn from the uniform distribution over $\orthg{d}$ (the set of $d \times d$ orthogonal matrices).

The uniform isometry $M_o$ helps unify various task specifications. However, the uniform distribution encodes \emph{no hierarchical or structural} bias, motivating our work's need for alternative distributions.

\subsection{Extending Trajectory by Composition}\label{ssec:extending-traj}
Since the uniform distribution carries no hierarchical or structural bias, a natural next step is to stack \( \ell \) stages of random orthogonal preprocessing before each update. Regard one orthogonal transform as one unit step. Placing \( \ell \) such steps before each state, action, next state update does not increase the number of environment interactions; the simulation horizon remains unchanged. Instead, the learner receives inputs that have passed through \( \ell \) preparatory steps at every time index, so from the model’s perspective, the sequence behaves as if it contained \( \ell+1 \) times as many updates. This simple composition is expected to provide longer-range temporal evidence to the learner and to reveal cues of hierarchical structure without modifying the environment. The formal mechanism used in the sequel is not an actual increase in the environment horizon, but the shared-prefix dependence induced by the joint law of the projection family $(\rep(w))_{w \in \words}$.

\subsection{Challenges in Altering the Random Projection's Distribution}\label{ssec:problem-statement}
Building on the compositional idea of \cref{ssec:extending-traj}, one might attempt to change the distribution of the random projection by replacing a single \(U\) with a product of \(\ell\) random orthogonal factors. Yet the \emph{Haar} property of the uniform distribution over \(\orthg{d}\) makes this ineffective: if \(U_1,\dots,U_\ell\) are independently drawn from the uniform distribution on \(\orthg{d}\), then the product \(U_\ell \cdots U_1\) has the same distribution as a single draw. Consequently, simply stacking uniform orthogonal matrices cannot inject hierarchical or structural bias. The question then becomes \textbf{how to replace the uniform distribution on \(\orthg{d}\) with one that can impose a meaningful hierarchical structure}.

%% file: 010_body.tex
\section{Free Random Projection}\label{sec:frp}
In addressing the key question in \cref{ssec:problem-statement}, we focus on \textbf{the joint behavior of words in random matrices}, where sequences of matrix multiplications are considered as \emph{words}. In the realm of free probability theory \citep{voiculescu1992free,mingo2017free}, initiated by D.\,Voiculescu, \textbf{the collective behavior of random matrices is modeled with an algebraic construct called a \emph{free group}}. Consequently, we employ free groups to circumvent the challenges outlined in \cref{ssec:problem-statement}, aiming to construct a projection family whose joint statistics reflect the prefix-tree structure of reduced words.

\subsection{Free Groups and Random Matrices}\label{ssec:free_group}
Before providing definitions, we first develop an intuition for free groups. A free group $\F_2$ with two generators can be understood as the set of all possible words formed by concatenating the generators $a$, $b$ and their inverses $a^{-1}$, $b^{-1}$, where we can simplify words only by canceling adjacent elements and their inverses (e.g., $aa^{-1} = e$, $a^{-1}a = e$). For instance, in $\F_2$, an element might be $aba^{-1}b^2a$, where such a reduced word representation in generators is unique. This uniqueness is what makes free groups \emph{free} - \textbf{there are no other relations or constraints between the generators}, which exhibit a tree structure.
Rigorously, the uniqueness and the free group are defined as follows.
\begin{definition}
Let $\genset=\{a_1, a_2, \dots, a_\ngen\}$ be a set of $n$ elements. \emph{The free group on $\genset$}, denoted by $\F_\genset$, is defined by the following universal property: For any group $G$ and any mapping $\rep_0:\genset\to G$, there exists a \textbf{unique} group homomorphism $\rep:\F_\genset\to G$ extending $\rep_0$, that is, $\rep(vw) =\rep(v)\rep(w) \text{ for all } v,w\in \F_\genset$ and $\rep(a) = \rep_0(a)$ for $a \in \genset$. We may simply denote the group as $\F_n$, since, by the universal property, it depends only on $n$ up to isomorphism.
\end{definition}
Because reduced words are unique, the Cayley graph of $\F_n$ is an infinite $2n$-regular tree. \cref{fig:overview}~(left) illustrates the case $n=2$. In this paper we use this tree as an indexing picture for words and shared prefixes, rather than as a literal geometric claim about the image of $\rep$ or the sequence of sampled environments.

For the random projection in RL, we focus on two matrix groups: \(G=\orthg{d}\), the group of \(d \times d\) orthogonal matrices, and its subgroup \(G=\permg{d}\), the symmetric group represented by permutation matrices. A matrix \(U \in \permg{d}\) if and only if there is a permutation \(\sigma\) of \(\{1,2,\dots,d\}\) such that \(U_{ij} = 1\) if \(\sigma(j) = i\), and \(U_{ij} = 0\) otherwise. %
Suppose we are given $U_1, U_2, \dots, U_\ngen$ independently sampled from $\Unif(\orthg{d})$ (resp.\,$\Unif(\permg{d})$). 
By the universal property, we define the matrix representation $\rep: \F_\ngen \to \orthg{d}$ (resp.\,$\rep: \F_\ngen \to \permg{d}$) by setting $\rep(a_i) = U_i$ for $i= 1, 2,\dots, \ngen$.\label{align:rep}
For example, $\rep(a_1a_3a_2a_4^{-1}) = U_1 U_3 U_2 U_4^{-1}$; the representation $\lambda$ substitutes matrices for the generators $a_1, \dots, a_n$.

A natural question is whether the matrix representation $\lambda$ preserves the tree structure of $\F_n$.
Free probability theory \citep{voiculescu1992free, mingo2017free} asserts that, in the high-dimensional limit (i.e., as $d\to\infty$),  the $\rep$ for each of $\orthg{d}$ and $\permg{d}$ approximately preserves the free group structure in various senses, including inner products and stronger criteria \citep{nica1993asymptotically,voiculescu1998strengthened,collins2003moments,collins2006integration, collins2014strong}. For the inner products,  we have 
\begin{align}
\lim_{d\to\infty} \E [\langle \rep(v), \rep(w) \rangle_{\orthg{d}}]
= \langle v,w \rangle_{\F_n}, \quad v,w \in \F_n, \label{align:inner-product}
\end{align}
where $\langle U,V \rangle_{\orthg{d}}=d^{-1}\Tr U^\tp V$, and $\langle v,w \rangle_{\F_n} = 1$ if $v=w$ and $0$ otherwise.
In \cref{align:inner-product}, the expectation is taken with respect to the randomness of the base matrices $U_1,\ldots,U_n$ used to define $\lambda$, sampled independently from $\mathrm{Unif}(\orthg{d})$ (or $\mathrm{Unif}(\permg{d})$ in the permutation case). Thus, for fixed words $v,w \in \F_n$, one first samples the base matrices, then forms the products $\lambda(v)$ and $\lambda(w)$, and finally averages the normalized trace inner product $d^{-1}\Tr(\lambda(v)^\tp \lambda(w))$. In particular, \cref{align:inner-product} is a statement about the joint law of the family $(\lambda(w))_{w \in \F_n}$, not merely about the marginal distribution of a single projection.

\Cref{align:inner-product} should therefore be read as a statement about the joint statistics of the family $(\rep(w))_{w\in \F_n}$. Distinct reduced words have asymptotically vanishing pairwise normalized-trace overlaps in expectation, yet the family is not sampled independently: if two words share a prefix of length $k$, then the corresponding products $\rep(w)$ share the same prefix product of the base matrices. This common-prefix product structure is the operative hierarchy used by FRP. It supplies a shared-prefix dependence structure without requiring any literal geometric realization of the Cayley graph on the sphere. The resulting representation-side effect is analyzed via the kernel in \cref{ssec:average-kernel-analysis} and tested in \cref{sec:pomdps}.

\subsection{Free Random Projection in Multi-Environment Meta-RL}
\begin{algorithm}[th]
\caption{Meta RL environment step with FRP}
\begin{algorithmic}[1]
\Require Distribution of environments $\rho_\env$, Agent action $a$ and Environment termination $\eoe$
\Require Distribution of words $\worddist$,  Matrix representation $\lambda: \F_n \to \orthg{d}$.
\Function{StepEnvironment}{$a$, $\eoe$}
\If{the environment terminated ($\eoe$)}
\State Sample random environment $E\sim \rho_\mathcal{E}$
\State Sample random word $w \sim \worddist$
\State Initialize random observation projection $M_o = \scaling T_2 \rep(w) T_1^E$
\State Initialize random action projection $M_a$ 
\State Reset $E$ to receive an initial observation $\obs$
\State Apply the random observation projection  to  $\obs' = M_o \obs$
\State Initialize $r = 0$ and $\eoe = 0$ 
\State \Return $\obs', r, \eoe$
\Else
\State Apply the projection $a' = M_a a$
\State Step $E$ using $a'$ to receive the next observation $\obs$, reward $r$, and done signal $\eoe$
\State Apply the projection $\obs' = M_o \obs$
\State \Return $\obs', r, \eoe$
\EndIf
\EndFunction
\end{algorithmic}
\label{alg:fra}
\end{algorithm}
Based on the concept in \cref{ssec:free_group}, we define \emph{free random projection} in multi-environment Meta RL, which is a pre-training stage of ICRL.

Let $\rho_\env$ be a distribution of environments. We denote by $d^E$ the dimension of observation for an environment $E$. Assuming that $d^E$ is bounded with respect to $\rho_\env$, we choose the hidden dimension $d$ to be $d \geq \sup_{E\sim \rho_\env} d^E $.
Denote by $T_1^E$ the $d \times d^E$ rectangular identity matrix. This corresponds to zero-padding observations from $E$. Let $\din \in \N$ be the desired dimension of the common observation space. Denote by $T_2$ the $\din \times d$  rectangular identity matrix. %(This corresponds to a projection if $\din < d$.) 
\begin{definition}[Free Random Projection]\label{defn:frp}
Let $w$ be a random word distributed according to a fixed probability distribution $\worddist$ on $\F_n$. We define the \emph{free random projection} (FRP, in short) as $\scaling T_2 \lambda(w) T_1^E$.    
\end{definition}

The hierarchy exploited by FRP is the prefix tree of words in $\F_n$. During a resampling period, the base matrices $U_1,\ldots,U_n$ are kept fixed, so if two sampled words share a prefix of length $k$, then their projections share the corresponding prefix product in $\lambda(w)$. In this sense, longer shared word prefixes translate into longer common prefixes in the corresponding matrix products. FRP therefore departs from standard RP at the level of joint statistics rather than through nonvanishing pairwise normalized-trace overlaps between distinct words. \cref{alg:fra} exposes this shared-prefix dependence by resampling a word at episode boundaries while keeping the base matrices fixed during trajectory collection. We do not assume that the sequence of environments itself forms a tree; rather, the tree serves only as an index set for a projection family with shared-prefix dependence, intended to bias the learner toward representations useful for tasks with branching latent futures.

We utilize the same action projection matrix $M_a$ as described in \citep{lu2023structured}. To ensure temporal consistency, a single episode utilizes the same word $w$ for all time steps within the same environment.

\subsection{Implementation Details}
\paragraph{Word Distribution}
We define the following collection of words to investigate the impact of word length for $n, \ell \in \N$:
\begin{align}
    \words := \{ a_{i_1} \dots a_{i_\length} \in \F_\ngen \mid i_1, \dots, i_\ell \in [\ngen] \}.\label{align:word-family}
\end{align}
%To ensure each word is distinct up to inverses, we do not include the inverses of generators.
We define the word distribution as $\worddist :=\Unif(\words)$, and sample words \(\{w_1,\dots,w_{\nenv}\}\) independently from $\worddist$ for the \(\nenv\) parallel environments.
In our analysis, we fix the total number of possible words $\nwords > \nenv$ and compare various pairs of $(\ngen, \length)$ that satisfy \(|\words|=\nwords\) to ensure a fair comparison. 
%In particular, writing $\lngen=\log_2 \ngen$, we have $\ngen^\length=2^{\lngen \length}=\nwords$. 

\paragraph{Resampling Period}
The matrix representation $\rep$ relies on the sampling of $U_1, \dots, U_n$. Consequently, we must determine the appropriate period for resampling. A reasonable approach, especially when alternating between trajectory collection (through environment steps) and parameter updates, is to resample the matrices at the beginning of each new trajectory collection phase. This methodology is employed in the experiments detailed in \cref{sec:pomdps}. %More generally, one may adjust the resampling period depending on the task difficulty and the sizes of $\nwords$ and $\nenv$. However, resampling too frequently negates the benefit of using a word distribution, effectively reverting to uniform random projection. Therefore, it is necessary to maintain the same set of random matrices for a certain number of trajectories before resampling.

\section{Kernel Analysis of FRP}\label{ssec:average-kernel-analysis}

\begin{figure*}[t]
    \centering
      \includegraphics[width=0.213\linewidth]{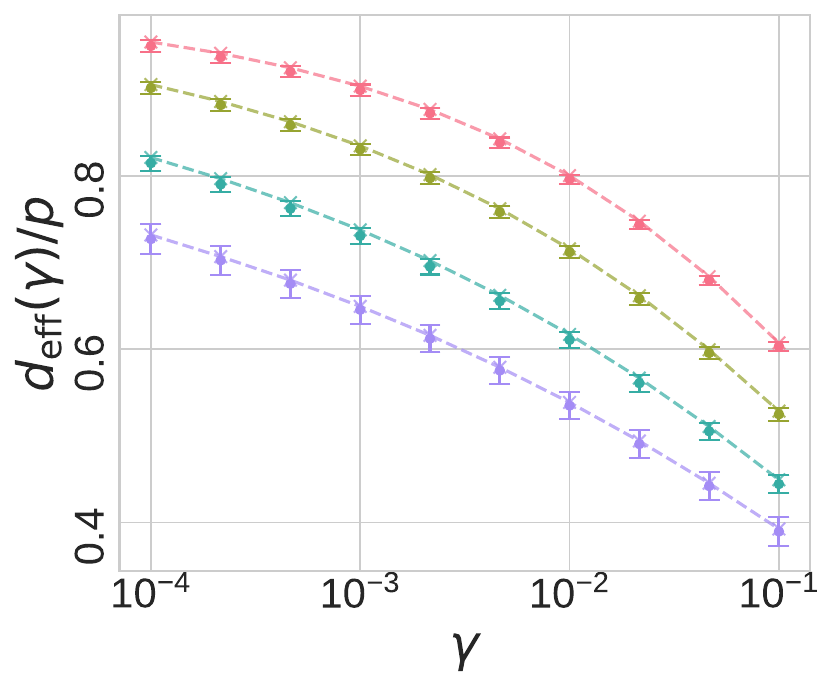}
    \includegraphics[width=0.78\linewidth]{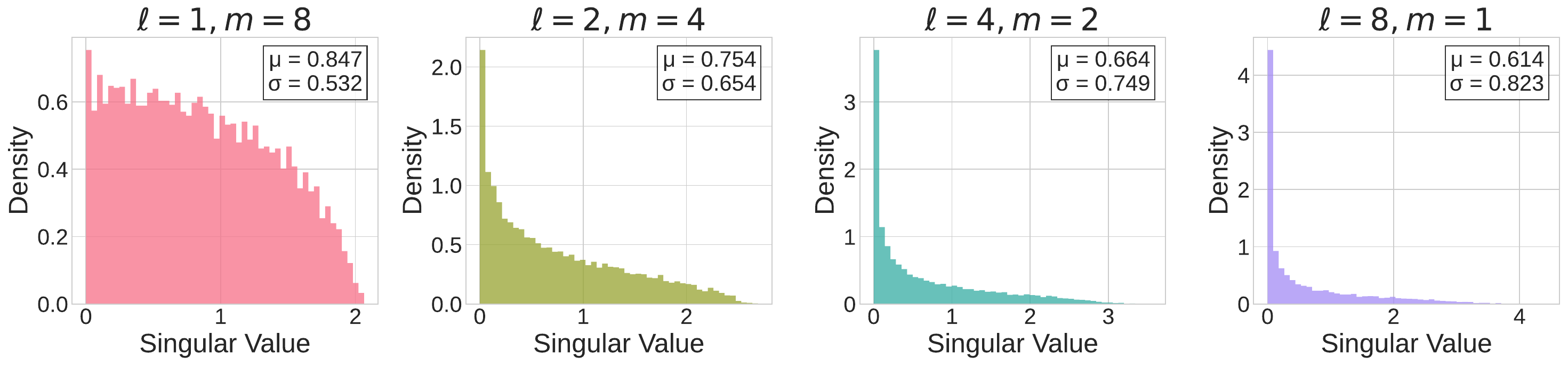}
    %\caption{ESD of $\nwords^{-1/2}\sum_{w \in \words}\lambda(w)$ for varying $\ell$ values (1, 2, 4, 8) with $m = \log_2 n= 8/\ell$. At \( \ell=1 \), FRP collapses to the standard RP. The parameters $d = 64$ and $\nwords =2^8$ are fixed. Each one is combined 128 trials, with a bin size of 50.}
    %\caption{Most Left: Theoretical (dotted) and empirical (error bars) $\effdim/p$ vs.\ $\ell\in\{1,2,4,8\}$. Rights: ESD of $\sqrt{\kermat}$ for \(\ell\in \{1,2,4,8\}\) with \(m=\log_2 n=8/\ell\). At \(\ell=1\), FRP collapses to the standard RP. Fixed \(d=64\), \(\nwords=2^8\). Histograms aggregate 128 trials (50 bins).}
    \caption{Left panel: theory (dotted) vs.\ empirics (error bars) of \(\effdim/p\) for \(\ell\in{1,2,4,8}\). Right four panels: ESDs of \(\sqrt{\kermat}\) for the same \(\ell\) with \(m=\log_2 n=8/\ell\) (\(\ell=1\) coincides with RP). Fixed (d=64), \(\nwords=2^{8}\). Histograms aggregate 128 trials (50 bins).}

    \label{fig:spectrum-average-frp}
\end{figure*}

This section analyzes the inductive bias introduced by the projection distribution itself. The object of study is not an RL-specific generalization bound, but the kernel induced by the projected inputs seen during meta-training. Standard random projection ($\ell = 1$) and FRP ($\ell > 1$) preserve the ambient dimension of each projected input, but they differ in the joint distribution of the projection family $(\lambda(w))_{w \in \words}$. Our goal is to quantify this difference spectrally.

In line with \citep{dao2019kernel}, we consider the averaged kernel matrix $\kermat$ for FRP, defined from samples $X_1,\ldots,X_p \in \R^d$. We then study its effective dimension as a representation-level complexity measure. In kernel methods, a smaller effective dimension is often associated with lower statistical complexity~\citep{MacKay1992, hsu2012random, zhang2015divide, tsigler2023benign}; here we use it more modestly, as a mechanistic summary of how FRP changes the geometry of the induced feature space. Accordingly, a decrease in effective dimension should be interpreted as evidence of a different inductive bias, rather than as a standalone guarantee of improved RL performance.

The averaged kernel matrix $\kermat$ and its \emph{effective dimension} are defined as
\begin{align}
\kermat &= \left(\frac{1}{\nwords}\sum_{w,w^\prime \in \words}\langle \lambda(w)X_i, \lambda(w^\prime)X_j \rangle\right)_{i,j=1}^{p}, \label{align:avg-kernel}\\
\effdim(\ell, \gamma)&=\Tr[\kermat(\kermat+ \gamma I_p)^{-1}], \quad \gamma>0. \label{align:eff-dim-def}
\end{align}
%Here, the scaling factor is not essential for the effective dimension, so we take beneficial one for the spectral analysis.
%
%

Based on $R$-diagonal elements \citep{nica2006lectures, haagerup2000brown}, we have the following estimation of $\kermat$'s effective dimension.
We briefly recall the spectral notation used in \cref{thm:eff-dim}. Let $X \in \R^{d \times p}$ be a data matrix whose columns are $X_1,\ldots,X_p$. For finite $d$ and $p$, the eigenvalue spectrum of $XX^\top$ is random; however, in the \emph{proportional regime} where both $d$ and $p$ are large with their ratio $p/d \to c$, the spectrum concentrates around a deterministic distribution. Formally, we consider a sequence of such matrices indexed by $d$, with $p = p(d)$ satisfying $p/d \to c \in (0,\infty)$, and assume that the empirical spectral distribution of $XX^\top$ converges in moments to a compactly supported measure $\nu$ on $[0,\infty)$. The resulting formula for the effective dimension (\cref{thm:eff-dim}) depends only on $\nu$ and provides an accurate approximation whenever $d$ and $p$ are moderately large (see \cref{sec:kernel-setup} for $d = p = 64$).

For a compactly supported probability measure $\mu$ on $[0,\infty)$ with first moment $m_1(\mu) = \int_\R t\,\mu(dt) \neq 0$, define the Cauchy transform
\begin{align}\label{align:cauchy}
G_\mu(z) = \int_\R \frac{1}{z - t}\,\mu(dt),
  \quad z \in \C \setminus \supp\mu,
\end{align}
and the $\psi$-function
\begin{align}\label{align:psi}
\psi_\mu(z) = \frac{G_\mu(1/z)}{z} - 1
  = \int_\R \frac{zt}{1 - zt}\,\mu(dt).
\end{align}
For $x < 0$ and $t \geq 0$ the integrand $xt/(1-xt)$ lies in $(-1,0)$, so $\psi_\mu$ maps $(-\infty,0)$ into $(-1,0)$. Moreover, $\psi_\mu'(x) = \int_\R t/(1-xt)^2\,\mu(dt) > 0$ for every $x \leq 0$, hence $\psi_\mu$ is a strictly increasing bijection $(-\infty,0) \xrightarrow{\;\sim\;} (-1,0)$. Its functional inverse $\chi_\mu\colon (-1,0) \to (-\infty,0)$ is therefore well-defined and analytic. The \emph{analytic $S$-transform} \citep{voiculescu1992free} is
\begin{align}\label{align:S-transform}
\mathscr{S}_\mu(z) = \frac{z+1}{z}\,\chi_\mu(z),
\end{align}
which extends analytically to a neighborhood of $(-1,0]$ (the apparent pole at $z=0$ is removable since $\mathscr{S}_\mu(0) = 1/m_1(\mu)$). Its key property is multiplicativity under free multiplicative convolution $\boxtimes$: $\mathscr{S}_{\mu \boxtimes \nu} = \mathscr{S}_\mu \cdot \mathscr{S}_\nu$.

The connection to the effective dimension is as follows. Let $\mu_{\kermat} = (1/p)\sum_{i=1}^p \delta_{\lambda_i}$ be the empirical spectral distribution of $\kermat$, where $\lambda_1 \geq \cdots \geq \lambda_p \geq 0$. Then
\begin{align}\label{align:eff-dim-psi}
\frac{\effdim(\ell,\gamma)}{p}
= \frac{1}{p}\Tr\!\left[\kermat(\kermat + \gamma I_p)^{-1}\right]
= -\psi_{\mu_{\kermat}}\!\Bigl(-\frac{1}{\gamma}\Bigr).
\end{align}
Hence the limit of $\effdim/p$ is determined by the limit of $\mu_{\kermat}$, which \cref{thm:eff-dim} computes via the $S$-transform. The proof is deferred to \cref{sec:proofs}.

\begin{thm}\label{thm:eff-dim}
    In the proportional regime $p/d \to c \in (0,\infty)$, assume that the empirical spectral distribution of $XX^\top$ converges to a compactly supported $\nu$ with $\int_\R t\,\nu(dt) \neq 0$.
    Fix $n,\ell \in \N$. Consider  $\lambda:\F_n \to \orthg{d}$ with \cref{align:rep}.
Then, for every $\gamma>0$, the limit $\tau(\ell,\gamma):=\lim_{p,d\to\infty,\,p/d\to c}\E[\effdim(\ell,\gamma)/p]$ exists, and $\tau(\ell, \gamma)= -\psi(-1/\gamma)$, where $\psi$ is the inverse function of $\chi$ defined by
    \begin{align}\label{align:chi-ftn-thm}
        \chi(z) = \frac{z}{z+1} \left(\frac{z/n + 1}{z + 1}\right)^\ell \mathscr{S}_\nu(z).
    \end{align}
Moreover, when comparisons across $\ell$ are made at fixed total number of words $\nwords = \ngen^\ell$ (equivalently, $\ngen = \nwords^{1/\ell}$), the quantity $\tau(\ell, \gamma)$ is strictly monotonically decreasing in $\ell$.
\end{thm}

In particular, $\tau(\ell, \gamma) \neq \tau(1, \gamma)$ ($\ell\neq 1$) by \cref{thm:eff-dim}, FRP's joint distribution with $\ell\neq1$ asymptotically departs  from that of the standard RP~(FRP with $\ell=1$).
\cref{fig:spectrum-average-frp}~(left)
%\cref{fig:effective-dim} 
compares the theoretical prediction $\tau(\ell,\gamma)$ for $\effdim(\ell, \gamma)/p$ from \cref{thm:eff-dim} with empirical results for finite dimension $p=d=64$ averaged over 128 trials; the theory aligns closely with the experimental observations. (See \cref{sec:kernel-setup} for the setup.) %Assuming $X_{ij}\sim\mathcal N(0,1/d)$ are i.i.d.\, and $c=1$,  
Notably, under the fixed-$\nwords$ comparison regime, for every $\gamma>0$, \textbf{larger $\ell$ consistently yields a lower effective dimension}, as predicted by \cref{thm:eff-dim}. We interpret this as evidence that FRP moves the induced kernel toward a lower-complexity representation regime; in the RL sections, this quantity is used as a mechanistic explanation rather than as a direct performance bound. Thus, although each $\lambda(w)$ maintains the data dimension in \cref{alg:fra}, their joint behavior implicitly reduces the effective dimension. This phenomenon results from the hierarchical structure represented by the degree-$\ell$ term $[(z/n+1)/(z+1)]^\ell$ in \cref{align:chi-ftn-thm}. 

%The spectral analysis of $\kermat$ as $d \to \infty$ is the proof's key.  
As shown in \cref{fig:spectrum-average-frp}~(right), increasing $\ell$ results in a heavier-tailed empirical singular value distribution (ESD) which in turn drives the observed reduction in effective dimension in \cref{fig:spectrum-average-frp}~(left). %\cref{fig:effective-dim} 
These spectral differences are also visible in block random matrices that probe higher-order joint statistics, as shown in \cref{ssec:kernel-analysis}.

\begin{comment}
\begin{wrapfigure}[13]{rh}{0.49\linewidth}
  \vspace{-\intextsep}
  %\vspace{0mm}
  \centering
  \includegraphics[width=\linewidth]{fig/norm_eff_dim_mp_d64_norm_t128.pdf}
  %\vspace{-8mm}
  \captionsetup{skip=2pt}
  %\caption{Theoretical (dotted lines) and empirical values (error bars) of $\effdim/p$ for varying $\ell$ in 1, 2, 4, 8.}
  \caption{Theoretical (dotted) and empirical (error bars) $\effdim/p$ vs.\ $\ell\in\{1,2,4,8\}$.}

  \label{fig:effective-dim}
  %\vspace{+0.5mm}
\end{wrapfigure}
\end{comment}

\section{Experiments}\label{sec:pomdps}
We evaluate how free random projection affects generalization in in-context learning. We employ POMDPs as our evaluation framework for two key reasons: (1) to maintain consistency with previous random projection studies \citep{Laskin2023context}, and (2) because POMDPs naturally require temporal information processing, a capability that benefits from the hierarchical bias induced by FRP. 

\begin{comment}

\begin{table*}[th]
%(8z,128,tiling)
\centering
%\caption{Test performance comparison across POPGym environments; Stateless Cartpole, Higher Lower, Minesweeper, Repeat First, and Repeat Previous. 
% Each value indicates the mean and standard error of ICL-Test MMER at the last step across 10 random seeds.}
 \caption{ICL-Test MMER at the final step across POPGym (Stateless Cartpole, Higher–Lower, Minesweeper, Repeat First, Repeat Previous), reported as mean($\pm$)std over 10 seeds.}

\label{tab:final-results-test}
\begin{tabular}{lccccc}
\toprule
Method & S.\,Cartpole & H.\,L. & M.\,Sweeper & R.\,First & R.\,Previsous \\
\midrule
RP(GRU) & $2.90\pm0.42$ & $0.18\pm0.29$ & $-1.31\pm0.50$ & $-4.88\pm0.47$ & $-7.66\pm0.28$ \\
RP(S5) & $2.87\pm0.29$ & $0.08\pm0.14$ & $-1.77\pm0.65$ & $-6.64\pm0.52$ & $-7.87\pm0.03$ \\
FRP(GRU) & $\mathbf{4.08\pm0.62}$ & $\mathbf{2.13\pm1.12}$ & $\mathbf{-0.69\pm0.56}$ & $\mathbf{-4.48\pm0.76}$ & $\mathbf{-6.04\pm1.65}$ \\
FRP(S5) & $3.85\pm0.55$ & $1.04\pm0.82$ & $-1.39\pm0.38$ & $-6.50\pm0.40$ & $-6.51\pm1.10$ \\
\bottomrule
\end{tabular}
\end{table*}
\end{comment}

\begin{table*}[ht]
\centering
\caption{Summary of metrics across tasks (ICL-Test MMER and raw values at the final epoch). Bold indicates the best among the four methods for each metric within each task. Reported as mean($\pm$)std, median, IQM over 10 seeds.}
\label{tab:final-results-test}
\begin{tabular}{llrrrrrr}
\hline
\multicolumn{1}{c}{Task} & \multicolumn{1}{c}{Method} & \multicolumn{3}{c}{Maximum Statistics (MMER)} & \multicolumn{3}{c}{Raw Statistics\ }\\
\cline{3-5}\cline{6-8}
 &  & Mean $\pm$ Std & Median & IQM & Mean $\pm$ Std & Median & IQM \\
\hline
\multirow{4}{*}{Stateless Cartpole}
 & RP(GRU)  & $2.90\pm0.42$ & $2.74$ & $2.79$ & $2.10\pm0.22$ & $2.03$ & $2.03$ \\
 & RP(S5)   & $2.87\pm0.29$ & $2.77$ & $2.78$ & $2.28\pm0.30$ & $2.18$ & $2.21$ \\
 & FRP(GRU) & {\bfseries\boldmath $4.08\pm0.62$} & {\bfseries\boldmath $4.05$} & {\bfseries\boldmath $4.16$} & {\bfseries\boldmath $3.63\pm0.73$} & {\bfseries\boldmath $3.58$} & {\bfseries\boldmath $3.68$} \\
 & FRP(S5)  & $3.85\pm0.55$ & $3.87$ & $3.92$ & $3.35\pm0.49$ & $3.55$ & $3.52$ \\
\hline
\multirow{4}{*}{Higher Lower}
 & RP(GRU)  & $0.18\pm0.29$ & $0.05$ & $0.06$ & $-1.84\pm1.03$ & $-2.37$ & $-2.32$ \\
 & RP(S5)   & $0.08\pm0.14$ & $0.05$ & $0.04$ & $-2.80\pm1.49$ & $-2.79$ & $-2.84$ \\
 & FRP(GRU) & {\bfseries\boldmath $2.13\pm1.12$} & {\bfseries\boldmath $1.88$} & {\bfseries\boldmath $2.03$} & {\bfseries\boldmath $0.65\pm1.50$} & {\bfseries\boldmath $0.78$} & {\bfseries\boldmath $0.62$} \\
 & FRP(S5)  & $1.04\pm0.82$ & $1.08$ & $1.05$ & $-0.21\pm0.86$ & $-0.31$ & $-0.30$ \\
\hline
\multirow{4}{*}{Minesweeper}
 & RP(GRU)  & $-1.31\pm0.50$ & $-1.45$ & $-1.44$ & $-1.64\pm0.70$ & $-1.67$ & $-1.63$ \\
 & RP(S5)   & $-1.77\pm0.65$ & $-1.69$ & $-1.75$ & $-2.06\pm0.69$ & $-2.02$ & $-2.05$ \\
 & FRP(GRU) & {\bfseries\boldmath $-0.69\pm0.56$} & {\bfseries\boldmath $-0.75$} & {\bfseries\boldmath $-0.75$} & {\bfseries\boldmath $-0.89\pm0.72$} & {\bfseries\boldmath $-0.98$} & {\bfseries\boldmath $-0.96$} \\
 & FRP(S5)  & $-1.39\pm0.38$ & $-1.37$ & $-1.37$ & $-1.63\pm0.51$ & $-1.56$ & $-1.58$ \\
\hline
\multirow{4}{*}{Repeat First}
 & RP(GRU)  & $-4.88\pm0.47$ & $-4.84$ & $-4.87$ & $-8.24\pm1.18$ & $-8.18$ & {\bfseries\boldmath $-8.05$} \\
 & RP(S5)   & $-6.64\pm0.52$ & $-6.55$ & $-6.58$ & {\bfseries\boldmath $-8.11\pm0.14$} & $-8.12$ & $-8.12$ \\
 & FRP(GRU) & {\bfseries\boldmath $-4.48\pm0.76$} & {\bfseries\boldmath $-4.73$} & {\bfseries\boldmath $-4.72$} & $-8.21\pm0.92$ & $-8.43$ & $-8.29$ \\
 & FRP(S5)  & $-6.50\pm0.40$ & $-6.63$ & $-6.63$ & {\bfseries\boldmath $-8.11\pm0.20$} & {\bfseries\boldmath $-8.11$} & $-8.12$ \\
\hline
\multirow{4}{*}{Repeat Previous}
 & RP(GRU)  & $-7.66\pm0.28$ & $-7.84$ & $-7.76$ & $-11.61\pm1.82$ & $-11.97$ & $-12.00$ \\
 & RP(S5)   & $-7.87\pm0.03$ & $-7.88$ & $-7.88$ & $-11.43\pm1.07$ & $-11.35$ & $-11.41$ \\
 & FRP(GRU) & {\bfseries\boldmath $-6.04\pm1.65$} & {\bfseries\boldmath $-6.32$} & {\bfseries\boldmath $-6.26$} & {\bfseries\boldmath $-6.81\pm2.07$} & {\bfseries\boldmath $-6.74$} & {\bfseries\boldmath $-6.71$} \\
 & FRP(S5)  & $-6.51\pm1.10$ & $-6.35$ & $-6.42$ & $-8.05\pm1.75$ & $-7.93$ & $-7.83$ \\
\hline
\end{tabular}
\end{table*}

\subsection{Evaluation Protocol}
We use the multi-environment training approach of \citet{lu2023structured}, alternating trajectory collection and parameter updates with standard proximal policy optimization (PPO) and generalized advantage estimation (GAE) \citep{schulman2015gae,schulman2017ppo}. FRP changes only the input projection; the optimizer and recurrent architectures are otherwise unchanged. Training schedules, trajectory-collection details, FRP hyperparameters, and architecture details are summarized in \cref{app:training-protocol,ssec:rps-hps,ssec:arch-hps}; \cref{eval-environment} specifies the deterministic evaluation projections used at test time, and \cref{ssec:ppo-loss} records only PPO-specific implementation choices.

Our protocol separates meta-training from test-time adaptation. During training, we jointly train a single recurrent agent on multiple environments sampled from the training set, using either FRP or standard RP to map heterogeneous observation and action spaces into a common input space. For evaluation, we hold out one environment type that is never used during training and fix a deterministic evaluation projection for that environment (the tiling or identity map described in \cref{eval-environment}), shared between RP and FRP. We then collect held-out trajectories with the same trajectory-collection hyperparameters as in \cref{tab:hps-traj}, but without any parameter updates. Each held-out episode consists of $16$ trials; the recurrent hidden state is reset at episode boundaries, and adaptation occurs only through hidden-state evolution within an episode. The reported ICL-Test MMER aggregates performance on these held-out episodes, so any RP--FRP difference reflects the inductive bias learned during meta-training rather than gradient-based adaptation at test time.

We adopt a recurrent model structure as in \citep{lu2023structured} that enables the in-context adaptation: $h_t, x_t = \RNN(h_{t-1},\obs'_t,\eoe_t)$, $\pi = \PolicyNet(x_t)$, $V = \ValueNet(x_t)$, where \(\RNN\) denotes either GRU (Gated Recurrent Unit) or S5 (Simplified Structured State Space Sequence Model)~\citep{lu2023structured, smith2023simplified} based recurrent module, \(\theta^r\) represents its learnable parameters, and \(\eoe_t\) is an indicator for episode termination. The hidden state \(h_t\) serves as the context window, capturing essential historical information that enables adaptation without parameter updates.

We conduct experiments using the pure-JAX reimplementation \citep{lu2023structured} of the Partially Observable Process Gym (POPGym) \citep{morad2023popgym} with components \citep{gymnax2022github}, a benchmark suite designed for memory-intensive RL tasks. \cref{app:implementation} summarizes the environment setup of POPGym (Stateless Cartpole, Higher Lower, Minesweeper, Repeat First, Repeat Previous).
POPGym modifies standard RL environments to remove direct access to certain state variables, making them partially observable, and testing the agent's ability to leverage historical information effectively.

\subsection{Generalization Gain}
\begin{figure}[th]
    \centering
    \includegraphics[trim=0 30 0 0, clip,width=\linewidth]{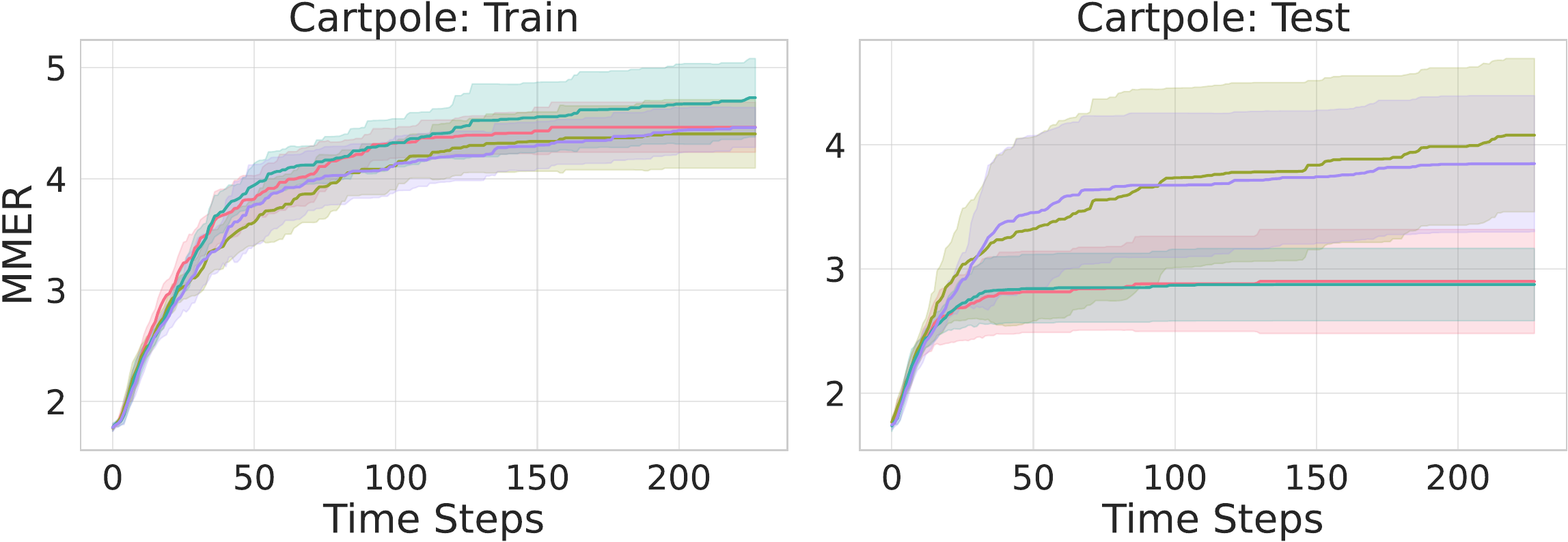}
    \includegraphics[width=\linewidth]{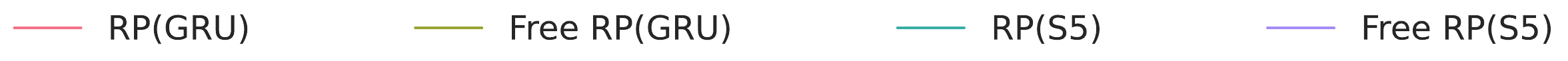}
    % dim 128
    %\caption{
    %Performance of FRP  vs. standard RP on four environments – Stateless Cartpole, Repeat Previous, Minesweeper, and Repeat First – shown in the top-left, top-right, bottom-left, and bottom-right subplots, respectively. We use $\ell^*$ in \cref{tab:optimal_depth} for FRP. Each subplot plots Train MMER and ICL-Test MMER. Shaded regions indicate standard error across 10 random seeds. }
    %\caption{
    %FRP  vs. standard RP on Stateless Cartpole. We use $\ell^*$ in \cref{tab:optimal_depth} for FRP. Each subplot plots Train MMER and ICL-Test MMER. Shaded regions indicate standard error across 10 random seeds. FRP was better than RP, over the standard error.}
    \caption{FRP vs.\ standard RP on Stateless Cartpole. FRP uses the optimal depth $\ell^*$ from \cref{tab:optimal_depth}. Each panel shows Train MMER and ICL-Test MMER curves; shaded bands: $\pm$ std\ over 10 seeds. FRP consistently outperforms RP beyond the error bands for both S5 and GRU.}

    %\label{fig:meta-envs-length-bestcase}
    \label{fig:cartpole-length-bestcase}

\end{figure}

\cref{tab:final-results-test}
%\cref{fig:meta-envs-length-bestcase} 
demonstrates improved ICL-test metrics (refer to \cref{ssec:eval}) when replacing standard random projection (baseline) with FRP across several POPGym environments.  We measure performance using the Max-Mean Episodic Return (MMER), defined as the highest average episodic return achieved by the agent across all training epochs, thereby capturing its peak performance.
In the Stateless Cartpole and Repeat Previous tasks, the benefits of  ICL eventually diminish under the standard random projection for both S5 and GRU. This observation suggests that FRP helps mitigate overfitting. Notably, in \cref{fig:cartpole-length-bestcase}, while RP with S5 achieves the highest training performance for Stateless Cartpole, FRP with GRU outperforms it during testing. This finding is particularly significant, as it illustrates how a relatively simpler model (GRU) combined with FRP's implicit hierarchical bias can outperform more structurally sophisticated models like S5.

%\begin{wraptable}{r}{0.48\linewidth}
\begin{table}[ht]
\centering
%(8z, 128, tiling)
\caption{Optimal word length ($\ell^*$)  for each environment and architecture. We see $\ell^*>1$ for all cases.}
  \captionsetup{skip=2pt}
\label{tab:optimal_depth}
\begin{tabular}{lccccc}
\toprule
Arch. & S.\,C. & H.\,L. & M.\,S. & R.\,F. & R.\,P. \\
\midrule
GRU & 4 & 8 & 4 & 2 & 2 \\
S5 & 4 & 4 & 8 & 2 & 4 \\
\bottomrule
\end{tabular}
%\vspace{-5mm}
\end{table}
%\end{wraptable}

%Moreover, in the Minesweeper tasks shown in \cref{fig:meta-envs-length-bestcase}, each model exhibits comparable training performance (FRP vs.\,standard); FRP consistently outperforms in testing. This observation underscores FRP's robust generalization capabilities across architectures.
As illustrated in \cref{tab:final-results-test}, we compare the final performance after training, with FRP(GRU) achieving the highest score. The consistency of improvement across diverse task types from continuous control (Stateless Cartpole) to memory-intensive tasks (Repeat Previous) indicates that the hierarchical bias introduced by FRP enhances the model's ability to capture temporal dependencies and adapt to new tasks without parameter updates, a key requirement for ICL.

\subsection{Impact of Word Length}

\cref{tab:optimal_depth} shows the optimal word length \(\ell^*\) in $\{1,2,4,8\}$ for each environment--architecture combination, determined by maximizing the test MMER. The optimal lengths vary across environments; in particular, \(\ell^* \geq 4\) in Stateless Cartpole, Higher Lower, and Minesweeper for both GRU and S5. This observation indicates that the benefit of strengthening the hierarchical bias depends on the specific environment.
Representative length-sweep curves are provided in \cref{ssec:impact-of-length}.

%\section{Analysis of Hierarchical Properties}
%In \cref{sec:frp}, we introduced FRP based on the concept of hierarchical bias and showed in \cref{sec:pomdps} that it improves performance in deep RL. To clarify the mechanism that yields these benefits, \com{TODO:fix here:we split the problem into two parts and conduct a detailed analysis}:
%(1) verifying, via a solvable model, that FRP indeed leverages the hierarchical structure of the state space, (2) FRP induces spectral bias when considering joint behavior by confirming that the effective dimension of FRP decreases with larger $\ell$,

\begin{figure*}[t]
    \centering
    \includegraphics[width=\linewidth]{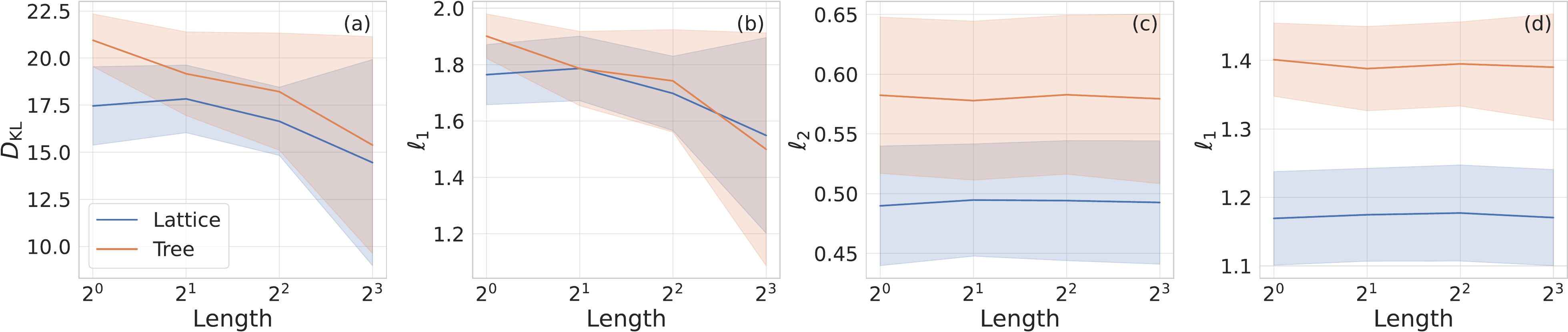}
    %\includegraphics[width=0.49\linewidth]{example-image-b}
    %\caption{
    %Effect of increasing word length $\ell$ on the meta-LSMDP performance metrics (lower values indicate better). 
    %We compare two types of state spaces: lattice (blue) and binary tree (orange). Panels show: (a)~KL-divergence $D_\mathrm{KL}(\pi^*\|\pi^\ell)$, (b)~$\ell_1$ policy error $\|\pi^\ell-\pi^*\|_1$,  (c)~$\ell_2$ and (d)~$\ell_1$ distance between $z^\ell/\|z^\ell\|_2$ and $ z^*/\|z^*\|_2$. Results show mean and std over 10 random seeds.
    %At \( \ell=1 \), FRP collapses to the standard RP.
%}
\caption{Effect of increasing $\ell$ on meta-LSMDP metrics (lower is better). Lattice (blue) vs.\ binary tree (orange). Panels: (a) $D_{\mathrm{KL}}(\pi^*|\pi^\ell)$; (b) $\|\pi^\ell-\pi^*\|_1$; (c) $\|z^\ell/|z^\ell\|_2 - z^*/\|z^*|_2\|_2$; (d) $\|z^\ell/\|z^\ell\|_2 - z^*/\|z^*\|_2\|_1$. Mean and std over 10 seeds. At $\ell=1$, FRP collapses to the standard RP.}
    \label{fig:meta-lsmdp}
\end{figure*}

\subsection{Linearly Solvable MDP}\label{ssec:lsmdp}
To  clarify the mechanism that yields FRP's implicit bias, we verify, via a solvable model, that FRP indeed leverages the hierarchical structure of the state space.
The linearly solvable Markov decision process (LSMDP) \citep{todorov2006linearly} provides analytic solutions for optimal policies. This analytical framework allows us to quantify the impact of FRP on generalization performance. In a typical MDP defined by the tuple $(\mathcal{S}, \mathcal{A}, P, R, \gamma)$, one solves for a policy that maximizes or minimizes long-term reward or cost through a Bellman equation, often requiring iterative methods. The LSMDP modifies this framework by introducing a passive transition $p(s'|s)$ and a control cost based on the KL divergence between the chosen policy $\pi(s'|s)$ and $p(s'|s)$. 
In an LSMDP, the total discounted cost includes both a state-based cost $c(s)$ and a KL cost $\alpha\,D_{\mathrm{KL}}(\pi \| p)$. A key insight is that the optimal policy $\pi^*$ is proportional to $p(\cdot|s)\exp\bigl(-(\gamma /\alpha)V^*(s')\bigr)$. This exponential form leads directly to a linear equation if we define the desirability function $z^*(s) = \exp\bigl(-(\gamma/\alpha)V^*(s)\bigr)$.
Concretely, $z^*(s)=\exp\bigl(-(\gamma/\alpha)\, c(s)\bigr)\sum_{s'} p(s'|s)\,z^*(s')$ and $\pi^*(s^\prime \mid s) = p(s^\prime | s)\, z^*(s^\prime)\big/\!\sum_{s''}p(s''\mid s)z^*(s'')$.
Due to Perron-Frobenius theory, a unique (up to scaling) positive solution $z^*$ exists when $\gamma\in(0,1)$.
%By solving for $z^*(s)$, one obtains $V^*(s) = -(\alpha/\gamma)\ln\,z^*(s)$ and the optimal policy $\pi^*$.
%LSMDPs thus allow analytical solutions and evaluation of the generalization performance.

%\subsection{Evaluation Protocol}
Then consider a meta-learning setting on the LSMDP. We may assume $\states=\{1,2,\dots,d\}$. To create parallel environments, we consider random permutations over $\states$.
Since permuting the indices of $\states$ does not change the essential nature of the problem, the optimal desirability function for each environment is given by $Uz^*$ for a $U \in \permg{d}$.  Therefore, similar to our previous approach in \cref{sec:pomdps}, we incorporate FRP in creating parallel environments by the matrix representation $\rep: \F_\ngen \to \permg{d}$.  We aggregate the solutions as $z^\ell = \nwords^{-1}\sum_{w \in \words} \rep(w)z^*$ and $\pi^\ell(s^\prime \mid s) = p(s^\prime | s)\, z^\ell(s^\prime)\big/\!\sum_{s''}p(s''\mid s)\,z^\ell(s'')$.
For comparison purposes, we consider the word generating sets $\words$ as described in \cref{align:word-family} with $\ell=1,2,4,8$.
We evaluated the generalization performance by measuring the KL divergence $D(\pi^* \| \pi^\ell)$.
Complete construction details for the lattice/tree state spaces, cost assignment, and FRP hyperparameters used in this meta-LSMDP study are given in \cref{sec:lsmdp-setup}.

\Cref{fig:meta-lsmdp} demonstrates improved generalization with increasing $\ell$ in both lattice and tree state spaces, with substantially larger gains on trees. Thus, the effect is not restricted to a single topology, but it is most pronounced when the state space is explicitly hierarchical.

Tree-structured state spaces demonstrated greater reductions in KL divergence compared to lattice structures, indicating FRP's high efficacy in hierarchical state spaces, as illustrated in \cref{fig:meta-lsmdp}~(a,b). While desirability-function metrics ($z$) remain relatively stable across varying word lengths in \cref{fig:meta-lsmdp}~(c,d), the policy metrics in \cref{fig:meta-lsmdp}~(a,b) improve more visibly. This suggests that, in this construction, changing $\ell$ affects the policy more strongly than the normalized desirability vector. We interpret this as evidence of a representation-side bias induced by FRP, while a sharper characterization of the responsible higher-order structure is left for future work.

%%%%%%%%%%%%%%%%%%%%%%%%%%%%%%%%%%%%%%%%%%%%%%%%%%%%%%%%%%%%

%% file: 070_discussion.tex
\section{Discussion}\label{sec:discussion}
Based on the hypothesis that hierarchical biases can benefit reinforcement learning, we introduce free random projection (FRP) for in-context reinforcement learning, using words in a free group to define a projection family with shared-prefix dependence. Notably, kernel analysis~(\cref{thm:eff-dim}) shows that FRP changes the joint distribution of the projection family and lowers the effective dimension $\effdim$. 
We empirically demonstrate that FRP enhances generalization performance across various partially observable MDP tasks in POPGym. To investigate the underlying reasons for this improvement, we analyze FRP within LSMDP and find that it better aligns with hierarchical structures in the state space.

Across all analyses (\cref{tab:final-results-test}, and \cref{fig:meta-lsmdp}), we observed that FRP ($\ell>1$) outperformed the standard random projection (RP), which corresponds to FRP with $\ell=1$.  However, in  POPGym study~\cref{tab:optimal_depth} the longest word length \( \ell \) was not always optimal; \( \ell^*\) depended on the environment. These observations indicate that improvements cannot be attributed solely to the reduction in effective dimension. This is expected: effective dimension characterizes properties of the projection-induced kernel, but it does not encode the agent–environment interaction that ultimately determines control performance. A natural next step is to develop generalization bounds for in-context and meta reinforcement learning with random projections that couple spectral properties of the projection  with properties of the underlying environment such as hierarchical structure. Such an analysis would clarify why moderate \( \ell>1 \) often suffices and how FRP’s implicit hierarchical bias plays a role.

%% file: 080_acknowledgement.tex
\section*{Acknowledgments}
T.\,H.\,was funded by Cluster, INC.\,and supported by JSPS Grant-in-Aid for Scientific Research (B) No.\,21H00987 and JST BOOST, Japan, Grant Number JPMJBY24G4.
B.\,C.\,was supported by JSPS Grant-in-Aid for Scientific Research (B) No.\,21H00987 and Challenging Research (Exploratory) No.\,23K17299.
T.\,H.\,acknowledges the hospitality of CIRM and ENS Lyon, where part of this project was conducted during B.\,C.'s tenure as Chaire Jean Morlet Holder.
We are grateful to Roland Speicher for accommodation support and insightful suggestions on block random matrices, and to Yuu Jinnai for helpful feedback on reinforcement learning.
We also thank the anonymous reviewers for their constructive comments, which helped improve the presentation of this paper.
Computational resources were provided by ABCI.

%% file: 090_appendix.tex
\section{Notations}
In this section, we summarize the notations commonly used throughout this paper. \cref{tab:notations-math,tab:notations-fra} provide lists of mathematical notations. In this study, we fix a probability space $(\Omega, \mathbb{P})$, denoting expectation as $\E$ and variance as $\mathrm{Var}$. In this paper, pairwise statistics refer to second-order quantities between two random variables, while higher-order joint statistics refer to moments involving multiple variables beyond the pairwise level.

\begin{table}[th]
    \centering
    \caption{Math Notations.}
    \begin{tabular}{cl}
    \toprule   
    \textbf{Symbol} & \textbf{Definition}\\
       \midrule
       $[n]$ & The set of $n$ integers $\{1,2,\dots, n\}$.\\
       $\N$ &   The set of non-negative integers.\\
       $\R$ &   The set of real numbers.\\      
      $\sphere{d}$ & The $d-1$ dimensional unit sphere $\{ \xi \in \R^d \mid \|\xi\|_2=1\}$.\\
      $\delta_{i,j}$ & $1$ if $i=j$ otherwise $0$.\\
      $\Tr(A)$ &  $\sum_{i=1}^d A_{ii}$ for $A \in \R^{d \times d}$. \\      
      $\|\xi\|_2$ & $(\sum_{i=1}^d \xi_i^2)^{1/2}$ for $\xi \in \R^d$.\\
      $\|\xi\|_1$ & $\sum_{i=1}^d |\xi_i|$ for $\xi \in \R^d$.\\
      $D_\mathrm{KL}(q\|p)$ & $-\sum_{i=1}^n (q_i \log p_i - q_i \log q_i)$. \\
      $\Unif(S)$ & The uniform probability distribution on a set $S$.\\ 
      \midrule
       $\mathrm{Aut}(S)$ & The group of all bijections from  $S$ to $S$.\\
       $\F_n$& The free group generated by $n$ generators.\\
       $\orthg{d}$ & \(\{U \in \R^{d \times d} \mid UU^\tp = U^\tp U = I_d\}\).\\
       $\permg{d}$ & \(\{U \in \orthg{d} \mid U_{ij}=\delta_{\sigma(i), j} (i,j \in [d]) \text{\ for  }\sigma \in \mathrm{Aut}([d]) \}\).\\
    \end{tabular}
    \label{tab:notations-math}
\end{table}

\begin{table}[th]
    \centering
    \caption{Notations of integers related to random projections.}
    \begin{tabular}{cl}
        \toprule
        \textbf{Symbol} &  \textbf{Definition}\\
        \midrule
       $\dobs$  & The dimension of the observation vector\\
       $\din$  & The dimension of the input vector of RNN\\
       
       $\drp$  & The dimension of random projection\\
       $\ell$ & Length of words \\
       $\ngen$  &  The number of generators \\
       $m$ & $\log_2 \ngen$\\
       $\nwords$ & The number of elements in word collections.\\
       %$\eoe$ & The indicator of the end of the episode\\
       
       \bottomrule
    \end{tabular}
    \label{tab:notations-fra}
\end{table}

Next, we summarize notations related to groups.
Here, a group is a set $G$ equipped with  $(\cdot, e)$, where $\cdot$ is a binary operation on $G$ and $e$ is the identity element in $G$. A group must satisfy the following four properties: closure, associativity, identity, and invertibility.
Let $G$ and $H$ be groups. A \emph{homomorphism} $\varphi \colon G \to H$ is a mapping such that for all $g_1, g_2 \in G$, $\varphi(g_1 g_2) = \varphi(g_1)\varphi(g_2).$ A homomorphism is called an \emph{isomorphism} if it is bijective. In that case, the groups $G$ and $H$ are said to be \emph{isomorphic}.
Then the representation of a group is defined as follows.
\begin{definition}[Representation and Action]\label{defn:rep}
Let $G$ be a group and $X$ be a set. Suppose we have a group homomorphism $\lambda \colon G \to \mathrm{Aut}(X)$,
where $\mathrm{Aut}(X)$ is the group of all bijections from $X$ to $X$. We call $\lambda$ a \emph{representation} of $G$ on $X$.
In this setting, we say that $G$ \emph{acts} on $X$ via $\lambda$, and write $G \curvearrowright X$.
\end{definition}

The term \textbf{representation} has a precise meaning in math: it refers to a homomorphism from an abstract algebraic structure (like a group as illustrated in~\cref{defn:rep}) into the group of linear transformations of a vector space, often realized concretely by matrices. In deep learning, however, representation usually denotes an internal feature encoding of data; for example, the output of a neural network's penultimate layer, which resides in what is often called a latent representation space.
In GRU-based deep reinforcement learning models, the term representation space typically refers to the space of the model's internal hidden states (the GRU's latent state), which encodes a compressed history of past observations and interactions. This latter notion of representation in deep neural networks is central in areas like contrastive learning and representation learning, where models learn useful feature embeddings from data. To avoid ambiguity between these two concepts, this paper implements a distinct naming convention: we designate \textbf{the group-theoretic concept as a matrix representation}, while referring to \textbf{the deep learning concept as a latent representation}.

\section{Cayley-Graph Visualization}\label{sec:hyperbolic}

We use the Cayley graph of $\F_n$ only as an intuitive visualization of the prefix-tree structure of reduced words. In particular, \cref{fig:overview}~(left) is a Poincar\'e-disk drawing of the Cayley graph of $\F_2$ and is included only for intuition. The technical arguments in the paper rely on the joint statistics of $(\rep(w))_{w\in\words}$ and the shared-prefix dependence they induce, not on a geometric realization of this graph on the sphere.

\begin{definition}[Cayley Graph]\label{defn:cayley}
The Cayley graph of a group $G$ with generating set $\genset$ is the graph whose vertices are the elements of $G$, with an edge from $g$ to $gx$ for each $g\in G$ and $x\in \genset$.
\end{definition}

\subsection{Drawing Rule for the Poincar\'e-Disk Visualization}
This section describes how the Cayley graph of $\F_2$ is drawn inside the Poincar\'e disk in \cref{fig:overview}~(left). As a set, the Poincar\'e disk is $\{x \in \R^2 \mid \| x \|_2 <1\}$, equipped with the hyperbolic distance
\begin{equation}
    d_H(x,y) = \mathrm{arcosh}\!\left(1 + 2\frac{\|x-y\|_2^2}{(1-\|x\|_2^2)(1-\|y\|_2^2)}\right).
\end{equation}
In this conformal model, geodesics are arcs of circles orthogonal to the boundary. The recursive construction below places the Cayley graph inside the disk so that each generator and its inverse correspond to a distinct family of orthogonal arcs (shown in different colors in \cref{fig:overview}):
\begin{enumerate}
\item Initialize $m=1$.
\item Define a base angle $\theta = \pi/m$ and a difference angle $\delta = \pi/(3m)$.
\item For each $k \in \{0, 1, \dots, 2m-1\}$, create points $x = (\cos(\theta k - \delta), \sin(\theta k - \delta))$ and $y = (\cos(\theta k + \delta), \sin(\theta k + \delta))$.
\item Draw the orthogonal circle passing through $x$ and $y$.
\item Recursively increase the density by multiplying $m$ by $3$ at each depth level.
\end{enumerate}
This procedure produces a tree that faithfully represents the prefix-tree structure of reduced words, with exponentially many branches converging toward the boundary.

A natural distance on the Cayley graph is the word metric:
\begin{definition}[Word Metric]
The word metric $d(g,h)$ between elements $g$ and $h$ of a free group is the length of the reduced word $g^{-1}h$, i.e., the minimum number of generators and their inverses needed to convert $g$ into $h$.
\end{definition}
Two words sharing a prefix of length $k$ are at word-metric distance $2(\ell - k)$ from each other, so the word metric directly reflects the depth at which their shared prefix ends. This is the geometric counterpart of the shared-prefix dependence structure exploited by FRP.

\section{Theoretical and Numerical Details on the Kernel}
\subsection{Proof of Theorem}\label{sec:proofs}

In this section, we provide the mathematical proof of \cref{thm:eff-dim}. To handle the sum $\sum_{w\in\words}\lambda(w)$ algebraically, we work in the group algebra $\C[\F_n]$ defined as follows. For spectral analysis, we use complex coefficients; by restricting the coefficients to $\R$ one obtains $\R[\F_n]$. For a comprehensive introduction to concepts related to $\C[\F_n]$, refer to \citep[Section~6.1]{mingo2017free}.

Let us denote by $\C[\F_n]$ the set of formal finite linear combinations of elements in $\F_n$ with complex coefficients,
$\alpha = \sum_{g \in \F_n} \alpha(g)g$, where only finitely many $\alpha(g) \neq 0$.
In other words, this is equivalent to assigning a complex number $\alpha_g$ to every element $g$ of the group. The product $\alpha \beta$ for $\alpha, \beta \in \C[\F_n]$ is defined by linearly extending the group product.
We define the adjoint operator $*$ on $\C[\F_n]$ by $(\sum_g \alpha_g g)^* = \sum_g \bar{\alpha}_g g^{-1}$.

Fix $n, \ell \in \N$ and define $T_n \in \C[\F_n]$ by
\begin{align}
    T_n = \sum_{i=1}^n a_i.
\end{align}
Then we have 
\begin{align}
\sum_{w \in \words}\lambda(w) = \lambda(T_n^\ell),
\end{align}
where $\lambda: \F_n \to \orthg{d}$ is linearly extended to $\lambda: \C[\F_n] \to M_d(\C)$.
Furthermore, the averaged kernel matrix $\kermat$ (\cref{align:avg-kernel}) satisfies
\begin{align}
    \kermat = X^\top \lambda\!\left(\frac{(T_n^*)^\ell T_n^\ell }{n^\ell } \right) X,
\end{align}
where $X=[X_1,\ldots,X_p] \in \R^{d \times p}$. (Here we use $\nwords=n^\ell$.)

Thus we define the spectral distribution of $(T_n^\ell)^* T_n^\ell$.
Define a sesquilinear form on $\C[\F_n]$ by setting
\begin{align}
\langle   \sum_g \alpha_g g,  \sum_h \beta_h h \rangle  = \sum_g \bar{\alpha}_g \beta_g.
\end{align}
From this we define $||a||_2 =  \sqrt{\langle a , a \rangle}$.
The completion of $\C[\F_n]$ with respect to $||\cdot ||_2$ consists of all functions $a : \F_n \to \C$ satisfying $\sum_{g \in \F_n} |a(g)|^2 < \infty$ and it is denoted by $\ell_2(\F_n)$ and is a Hilbert space. Denote by $B(\ell_2(\F_n))$ the algebra of  bounded linear operators on $\ell_2(\F_n)$. Here, a linear operator $A$ is said to be bounded if  $\sup\{ ||A\xi||_2 \mid \xi \in \ell_2(\F_n), ||\xi||_2=1 \} < \infty$. The adjoint $A^*$ is defined as the bounded operator uniquely determined by $\langle A^* \xi, \zeta \rangle = \langle \xi, A \zeta \rangle$ for any $\zeta, \xi \in \ell_2(\F_n)$.
Then  we can consider the inclusion $\C[\F_n] \subset B(\ell_2(\F_n))$ by $g \cdot \sum_h {\alpha_h} h = \sum_{h \in \F_n} \alpha_h gh$.   This is a $*$-homomorphism, that is, the product and the adjoint are consistent with those of $B(\ell_2(\F_n))$.
We define a linear functional $\tau$ on $\C[\F_n]$ by $\tau(w) = \langle e, w \cdot e \rangle=\delta_{e,w}$, where $e$ is the identity element of $\F_n$. The $\tau$ is a tracial state since $\tau(e)=1$, $\tau(aa^*)\geq 0$, and $\tau(ab)=\tau(ba)$ for $a,b\in \C[\F_n]$. 
Consider $a \in \C[\F_n]$ with $a=a^*$. The spectral distribution of $a$ is defined as the unique compactly supported probability distribution $\mu_a$ satisfying 
\begin{align}
     \tau(a^k) = \int_\R x^k \mu_a(dx), \quad k \in \N.
\end{align}
The distribution is compactly supported because $\C[\F_n] \subset B(\ell_2(\F_n))$.
We also have the support of $\mu_{aa^*}$ is contained in $[0,\infty)$.

We use the Cauchy transform $G_\mu$, the $\psi$-function, the analytic $S$-transform $\mathscr{S}_\mu$, and the effective-dimension identity $\effdim/p = -\psi_{\mu_{\kermat}}(-1/\gamma)$ introduced in \cref{ssec:average-kernel-analysis} (\cref{align:cauchy}--\cref{align:eff-dim-psi}). For $x \leq 0$, the derivative satisfies
\begin{align}
\psi_\mu'(x) = \int_\R \frac{t}{(1-xt)^2}\,\mu(dt) \geq \frac{m_1(\mu)}{(1+|x|\,\|\mu\|)^2} > 0,
\end{align}
where $\|\mu\| = \max\supp\mu$; in particular $\chi_\mu' > 0$ on $(-1,0)$.
Refer to \citep{mingo2017free} for the definitions of ($*$-)freeness, asymptotic freeness, free multiplicative convolution, Haar unitaries, and $R$-diagonal elements.
In particular, the generators $a_1, \dots, a_n$ of $\F_n$ are $*$-free Haar unitaries by definition and therefore $R$-diagonal. By direct computation, we have $\tau(T_n^*T_n) \neq 0$.
Now we have prepared to prove \cref{thm:eff-dim}.

\begin{lemma}\label{lem:eff-dim}
    Fix $n,\ell \in \N$. Consider  $\lambda:\F_n \to \orthg{d}$ with \cref{align:rep}.
    In the proportional regime $p/d \to c \in (0,\infty)$, assume that $X$ is independent of $(\lambda(a_1),\dots,\lambda(a_n))$ and that the empirical spectral distribution of $XX^\top$ converges to a compactly supported $\nu$ with $\int_\R t\,\nu(dt) \neq 0$.
    Then, it holds that
    \begin{align}
        \lim_{p,d\to\infty,\, p/d\to c}        \E\!\left[\frac{\effdim(\ell, \gamma)}{p}\right]=-\psi\!\Bigl(-\frac{1}{\gamma}\Bigr), \quad \gamma >0.
    \end{align}
     where $\psi$ is the inverse function of $\chi$ given by
    \begin{align}
        \chi(z) = \frac{z}{z+1} (\frac{z/n + 1}{ z + 1})^\ell \mathscr{S}_\nu(z).
    \end{align}
\end{lemma}
\begin{proof}
    By the assumption, as $p/d \to c$,
\begin{align}
    \E \!\left[\frac{1}{d}\Tr\!\left((XX^\top)^k\right)\right] \to \int_\R t^k \nu(dt), \quad k \in \N.
\end{align}
Write $\mu=\mu_{(T_n^*)^\ell T_n^\ell/n^\ell}$.
By the asymptotic freeness of Haar orthogonal matrices \citep{collins2003moments}, the following family is asymptotically free as the limit of $d$ and $p$:
\begin{align}
(\lambda(a_1),\lambda(a_1)^*),  \dots, (\lambda(a_n),\lambda(a_n)^*),  XX^\top.
\end{align}
Thus it holds that
\begin{align}
\E[    \int_\R f(t) \mu_{\kermat}(dt) ]\to \int_\R f(t) \mu \boxtimes \nu (dt),
\end{align}
for any bounded continuous function $f$.
Since $t(t+\gamma)^{-1}\leq 1$ for $\gamma>0$ and $t\geq 0$, we have 
\begin{align}
    \E\!\left[\frac{\effdim(\ell,\gamma)}{p}\right] = -\E\!\left[\psi_{\mu_{\kermat}}\!\Bigl(-\frac{1}{\gamma}\Bigr)\right] \to -\psi_{\mu \boxtimes \nu}\!\Bigl(-\frac{1}{\gamma}\Bigr),
\end{align}
as $d,p \to \infty$ with $p/d \to c \in (0, \infty)$.
Therefore, the remaining part of the proof is computing the $S$-transform.

By \citep[Example~5.5]{haagerup2000brown}, 
\begin{align}
       \mathscr{S}_{\mu_{T_n^* T_n}}(z) &= \frac{z+n}{n^2(z+1)},\\
       \mathscr{S}_{\mu_{T_n^*T_n/n}}(z) &= \frac{z/n+1}{z+1}.
\end{align}
As the $R$-diagonal property is preserved under addition~\citep[Proposition~3.5]{haagerup2000brown}, $T_n$ is therefore $R$-diagonal. Then, by its multiplicative property~\citep[Proposition~3.10(i)]{haagerup2000brown}, the spectral distribution of $(T_n^*)^\ell T_n^\ell$ is equal to that of the $\ell$-fold free multiplicative product of $T_n^* T_n$:
\begin{align}\label{align:prod-R-diag}
    \mu_{(T_n^*)^\ell T_n^\ell} = (\mu_{{T_n^*} T_n})^{\boxtimes \ell}.
\end{align}
The $S$-transform for the free multiplicative convolution is the product of each individual $S$-transform, hence \begin{align}
 \mathscr{S}_\mu(z) = \mathscr{S}_{\mu_{T_n^*T_n}/n}(z)^\ell = (\frac{z/n + 1 }{ z+ 1})^\ell.
\end{align}
Furthermore, $\mathscr{S}_{\mu \boxtimes \nu}(z) = \mathscr{S}_{\mu}(z)\mathscr{S}_{\nu}(z)$. 
\begin{align}\label{align:chi-origin}
    \chi_{\mu \boxtimes \nu}(z) = \frac{z}{z+1} \mathscr{S}_\mu(z) \mathscr{S}_\nu(z) = \frac{z}{z+1} (\frac{z/n + 1}{ z+ 1})^\ell \mathscr{S}_\nu(z).
\end{align}
Thus, by defining $\chi=\chi_{\mu \boxtimes \nu}$ and $\psi=\psi_{\mu \boxtimes\nu}$, the proof is completed.
\end{proof}

Based on \cref{lem:eff-dim}, we show that the normalized effective dimension decreases as $\ell$ grows in \cref{lem:mono}. Note that for evaluating FRP across different $\ell$, we maintained a constant total number of words, $\nwords$.
\begin{lemma}\label{lem:mono}
    Consider $\nwords, \ell \in \N$ and define $n=\nwords^{1/\ell}$. Under the condition of  \cref{lem:eff-dim}, write the limit normalized effective dimension by
\begin{align}
    \tau(\ell, \gamma) = \lim_{d,p} \E\!\left[\frac{\effdim(\ell,\gamma)}{p}\right] = -\psi_{\mu \boxtimes \nu}\!\Bigl(-\frac{1}{\gamma}\Bigr).
\end{align}
Then, for any $\gamma>0$, $\tau(\ell, \gamma)$ is a strictly monotonically decreasing function with respect to $\ell$.
\end{lemma}

\begin{proof}
    By \cref{lem:eff-dim}, we only need to focus on $\chi_{\mu \boxtimes \nu}$ and $\psi_{\mu \boxtimes \nu}$.
    Consider $\ell \in [1, +\infty)$ by extending the definition of \cref{align:chi-origin}. For $y \in (0, 1)$ and $\ell \in [1, +\infty)$, set 
    \begin{align}
        \mathcal{F}(\ell, y) = -\chi(-y)=\frac{y}{1-y}\mathscr{S}_\nu(-y)\left( \frac{1-y \nwords^{-1/\ell} }{ 1- y} \right)^\ell.
    \end{align}
    Since $\mathscr{S}_\nu$ is analytic on a neighbourhood of $(-\infty, 0]$, $\mathcal{F}(y, \ell)$ is analytic with respect to $y$.
    Now by direct calculation, 
    \begin{align}\label{align:part-y}
        \partial_y \mathcal{F} = - \chi'(-y) (-1)= \chi^\prime(-y)>0.
    \end{align}
    Further, 
    \begin{align}\label{align:part-ell}
        \partial_\ell \mathcal{F} = \frac{y}{1-y}\mathscr{S}_\nu (-y)\left( \frac{1- y \nwords^{-1/\ell} }{ 1- y} \right)^\ell \left[ \log \frac{1- y \nwords^{-1/\ell}}{1-y} + \frac{y\nwords^{-1/\ell} (\log \nwords)/\ell  }{1- y\nwords^{-1/\ell}} \right] > 0.
    \end{align}
    Therefore we can extend the definition of $\tau(\gamma, \cdot)$ on $\ell \in [1,+\infty)$ and 
    \begin{align}
        \frac{1}{\gamma} = \mathcal{F}(\ell, \tau(\ell, \gamma)).
    \end{align}
    Taking derivative with $\ell$ of both hand sides and using \cref{align:part-ell,align:part-y}, we have
    \begin{align}
        \partial_\ell \tau(\ell, \gamma) = - \frac{\partial_\ell \mathcal{F}}{\partial_y \mathcal{F}} <0.
    \end{align}
 Given the constraint $\ell \in \N$, the proof is completed. 
 \end{proof}

\begin{proof}[Proof of \cref{thm:eff-dim}]
The assertion directly follows from \cref{lem:eff-dim,lem:mono}.    
\end{proof}

\subsection{Implementation Details on Numerical Computations}\label{sec:kernel-setup}

This section is for the detailed implementation of experiments in \cref{ssec:average-kernel-analysis}, including computing the effective dimension and inverting algorithms of the $\psi$.

Throughout the experiments, we assume that $d=p=64$ and $X$ are sampled from i.i.d.\,$N(0,1/d)$. Further, we assume that $X$ is independent from $\lambda(w) (w \in \words)$. In this case, the limit spectral distribution $\nu$ with $p=d \to \infty$ is given by the Marchenko-Pastur distribution with the parameter $c=p/d=1$ and we get 
\begin{align}
    \mathscr{S}_\nu(z) = \frac{1}{z+c} = \frac{1}{z+1}.
\end{align}

We choose $\gamma>0$ from $[10^{-4}, 10^{-1}]$.
We sample 128 independently of $X$ and $\lambda(w) (w \in \words)$ and then compute $\kermat$ for each. Then compute its eigenvalues $\lambda_1 \geq \dots \geq \lambda_p \geq 0$.
For the empirical value of the $\E[\effdim(\ell,\gamma)/p]$, we compute the average of $ (1/p)\sum_{i=1}^p \lambda_i(\lambda_i+\gamma)^{-1}$ over trials.

To compute the theoretical value $-\psi(-1/\gamma)$, consider the following function based on $\chi(-y)$:
\begin{align}\label{align:newton-taget}
      F(y) = -\gamma y(-y/n+1)^\ell + (-y+1)^{\ell+1}(-y+c), \quad y \in (0,1),
\end{align}
where $c=1$.
Then the solution $y^*$ of $F(y)=0$ is the solution of $\chi(-y)=-1/\gamma$ and thus $y^*=-\psi(-1/\gamma)$.

We solve \cref{align:newton-taget} by the Newton method: $y_{t+1} = y_t - F(y_t)/F^\prime(y_t)$ starting from the initial point $y_0=0.5$ with the tolerance $10^{-7}$ for the convergence. We implement the Newton method with Python and Jax with computing the $F'$ with Jax's autograd. We observed all experiments finished within $t<1000$ iterations.

\subsection{Higher-order Joint Statistics and Word Length}\label{ssec:kernel-analysis}

\begin{figure}[th]
    \centering
    \includegraphics[width=\linewidth]{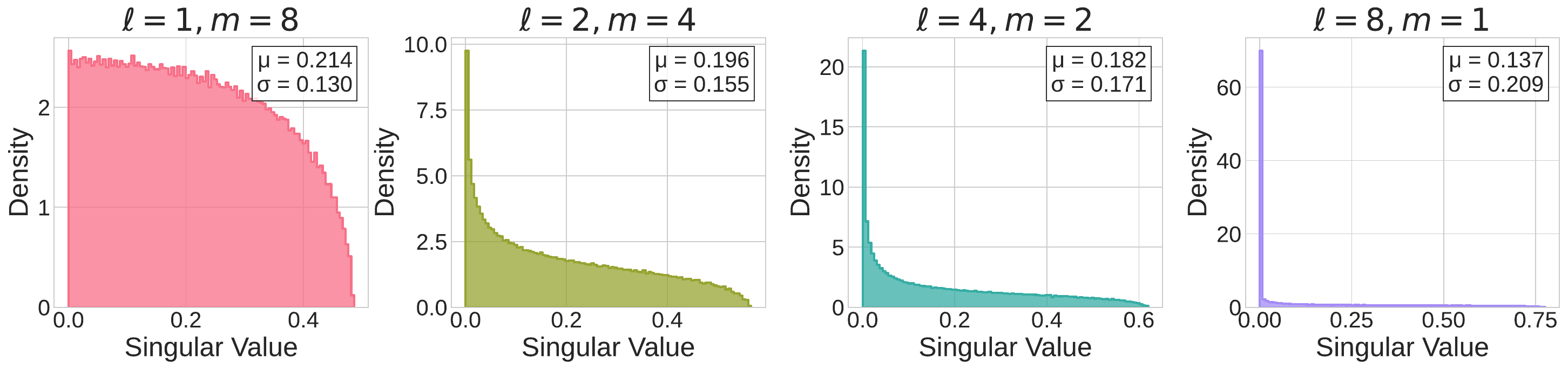}
    \caption{%\textbf{Differences in higher-order joint statistics through block-matrix spectrum}.
    Different lengths (\(\ell=1,2,4,8\)) exhibit distinct spectra of $K$. Each histogram shows the empirical distribution obtained 
    from 32 trials, which yields total \(32 d  \nwords^{1/2}=2^{15}\) values.}
    \label{fig:partial_trans}
\end{figure}

This section examines how the word distribution of FRP changes at the level of higher-order joint statistics. Using a block-matrix technique, we probe dependence patterns that are invisible to pairwise entry overlaps $\E[\langle \rep(w_i), \rep(w_j)\rangle]$ (which are asymptotically zero in expectation for $i \neq j$ by \cref{align:inner-product}). This avoids a direct analysis of numerous joint entry interactions.

For simplicity, assume \(\nwords = 2^{2k}\) for $k \in \N$. Let $\ell$ be a divisor of $2k$, and write $m=2k/\ell$ and $n=2^m$. Then, the elements in $\words$ have the form $a_{i_1} \dots a_{i_\ell}$ ($i_1, \dots, i_\ell \in [2^m]$). For comparison along distinct $\ell$, we decompose each component into $m$ components in $[2]$ and multi-index the elements as $(j_1,j_2, \dots, j_{2k})$: $w_{j_1, \dots, j_{2k}}$ (where $j_\mu \in [2]$ for $\mu \in [2k]$). We then arrange the components into a $2^k \times 2^k$ block matrix, using  $k$ indices as rows and the other $k$ indices as columns.

For a specific example, let $k=4$ and $\nwords=2^8$. Utilizing the partial transpose approach (refer to \cref{sec:ppt}), we derive a $2^4 \times 2^4$ matrix $A$, defined by the following index rearrangement:
\begin{align}
    A_{(j_1, j_7, j_3, j_5), (j_4, j_6, j_2, j_8)} = w_{j_1,j_2, \dots, j_8}, \text{ where } j_1, j_2, \dots, j_8 \in [2].\label{align:matA}
\end{align}
We compare the eigenvalue distributions of the $d\nwords^{1/2}  \times d\nwords^{1/2}$ random matrix $K = \nwords^{-1/2}\rep(A)\rep(A)^\tp$ across $\ell=1,2,4,8$, where $\rep(A)$ is the matrix obtained by applying $\lambda$ to matrix entries. As shown in \cref{fig:partial_trans}, for \(\ell=1\), the empirical eigenvalue distribution approximates the Marchenko-Pastur distribution. This distribution describes the asymptotic density of singular values for large random matrices with i.i.d. entries.
This behavior indicates that larger $\ell$ changes higher-order joint statistics involving simultaneous interactions among multiple entries, such as a linear constraint across an entire row, column, or block, rather than merely pairwise entry overlaps.

Finally, \cref{fig:partial_trans} shows that despite the \(d=64\) dimension of each representation matrix \(\lambda(w_i)\), the block random matrix \(Z\) exhibits distinct spectral characteristics. Thus, extremely high-dimensional matrices are unnecessary. By analyzing the joint word distribution with a sufficiently large number $\nwords$ of word patterns $\words$, the inherent hierarchical properties of free groups are preserved and reflected in the hierarchical bias of FRP.

\subsection{Partial Transpose}\label{sec:ppt}
This section investigates the role of the partial transpose, as described in \cref{ssec:kernel-analysis}. We aim to construct a block matrix using the following approach. Let $m$ be an even integer and consider the free group $\F_{2^m}$, generated by elements $a_1, a_2, \dots, a_{2^m}$, with words of length $\ell$ ordered lexicographically. These words are indexed in binary as $w_{i_1, \dots, i_{ml}}$, where each $i_\mu$ belongs to the set $[2]$. For example, when $m=2$ and $\ell=2$, the index is denoted as $w_{1,1}=a_{1}^2$, $w_{1,2}=a_1a_2$, $w_{2,1}=a_2a_1$, and $w_{2,2}=a_2^2$. Subsequently, we define $k=ml/2$ and $2^{k} \times 2^{k}$ matrix $W$ as follows:
\[
W_{(i_1, \dots, i_k), (j_1, \dots, j_k)} := w_{i_1, \dots, i_k, j_1, \dots, j_k}
\]
by arranging the elements accordingly.

In this basic arrangement of components, for $\ell>1$, the setup allows for a trivial decomposition: $W_{\mathbf{i},\mathbf{j}} = v_\mathbf{i}v_{\mathbf{j}}$ (where $\mathbf{i}, \mathbf{j} \in [2]^k$) for some vector $v$ containing words. This decomposition results in an eigenvalue degeneracy that masks distinct differences. 

\begin{figure}[th]
    \centering
    \includegraphics[width=\linewidth]{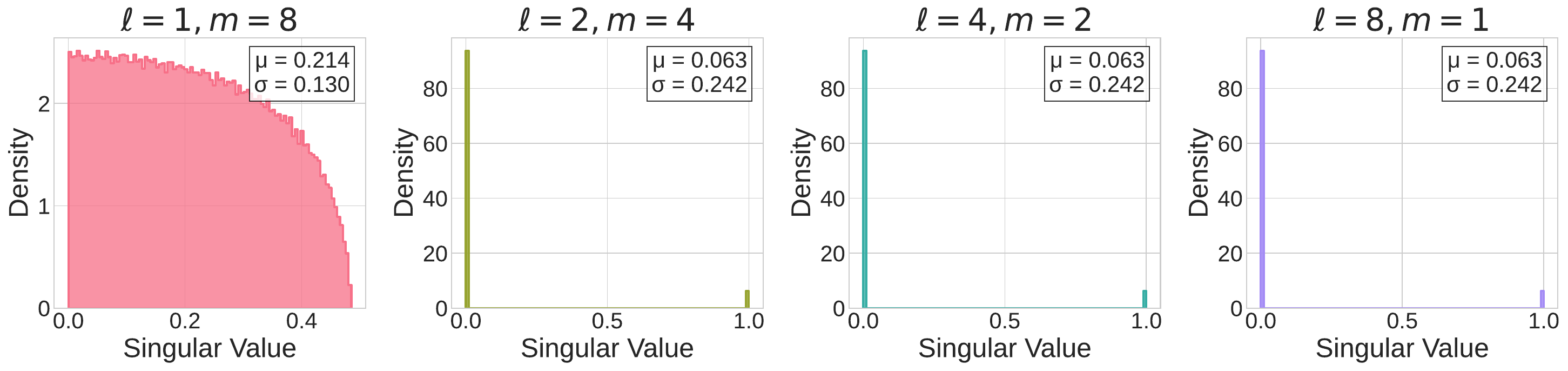}
    \includegraphics[width=\linewidth]{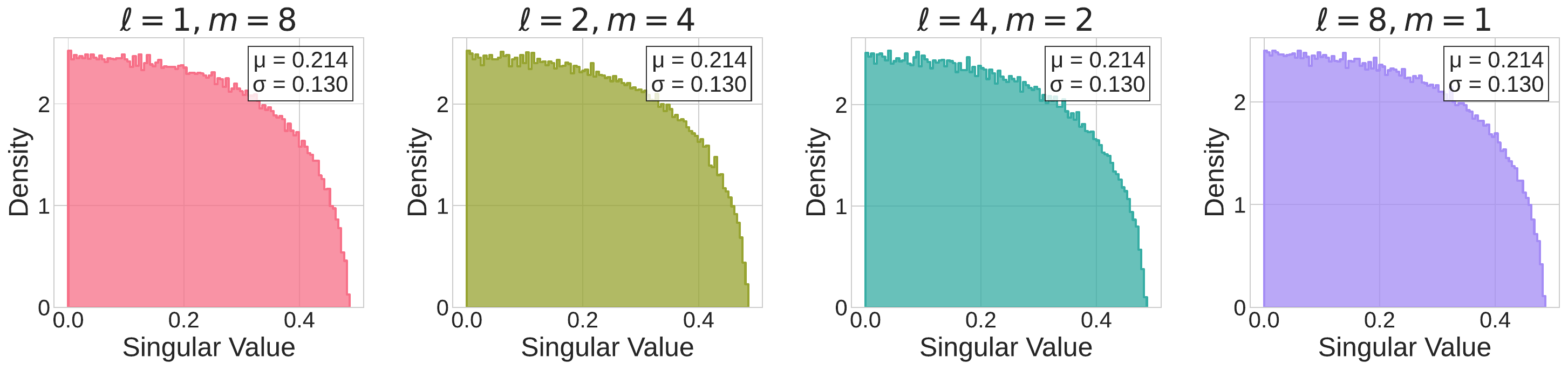}
    \caption{\textbf{Insufficiently distinctive} empirical spectrum of block random kernels with $\rep: \F_{2^m} \rightarrow \orthg{64}$. (Top) Raw block matrix without the partial transpose.
    (Bottom) After uniform shuffling entries of block random matrices.  Each histogram is based on 32 trials.}

    \label{fig:eigs-notrans}
\end{figure}

In particular, consider $\nwords=2^8$, $k=4$, and $(\ell,\log_2n)=(1,8), (2,4), (4,2), (8,1)$. In lexicographic order, the generator is arranged into a vector to create a vector of words $v$.
\begin{align}
    v = \begin{cases}
       &\{ a_1, a_2, \dots, a_{16}\}, \text{\ if\ } \ell=2,\\
       &\{ a_1^2, a_1a_2, \dots, a_4^2\}, \text{\ if\ } \ell=4,\\
       &\{ a_1^4, a_1^3a_2, \dots, a_2^4 \}, \text{\ if\ } \ell=8,\\
    \end{cases}
\end{align}
where $|v|=2^k$ holds for $\ell=2,4,8$.
In this setting, we decompose $W= vv^\tp$, making $W$ a rank-1 orthogonal projection with eigenvalues of either 1 or 0. Indeed, \cref{fig:eigs-notrans}~(Top) directly corresponds to the singular values of the block random matrix $\lambda(W)/\sqrt{2^k}$; it illustrates that for $\ell\geq2$, distinctions are not apparent. This indicates that a modification to $W$ is necessary.

In a straightforward modification approach, shuffling block matrix elements with the uniform distribution over $\permg{\nwords}$ results in identical distributions for all $\ell$, as shown in \cref{fig:eigs-notrans}~(Bottom). Thus, alternative methods are necessary to transform $W$.

To better differentiate these distributions, we apply a \emph{partial transpose} operation to the block structure of $W$, altering the component arrangement. The variations in singular values will still indicate differences in higher-order joint statistics, independent of the arrangement.
For a concrete case, consider $k=4, \nwords=2^8$, and define the partial transpose as a mapping between block matrices:
\[   
T_{(2,7)(4,5)}: W_{(i_1, i_2, i_3, i_4), (j_1, j_2, j_3, j_4)} \mapsto W_{(i_1, j_3, i_3, j_1), (i_4, j_2, i_2, j_4)}
\]
Then we have 
\[
A =  T_{(2,7)(4,5)}(W),
\]
where $A$ is the matrix defined in \cref{ssec:kernel-analysis}. As demonstrated in \cref{fig:partial_trans}, there are distinctly different empirical spectral distributions for different $\ell$.
Consequently, the partial transpose contributed to discerning statistical differences among the distributions we consider via $\ell$.

\begin{figure}[th]
    \centering
    \includegraphics[width=\linewidth]{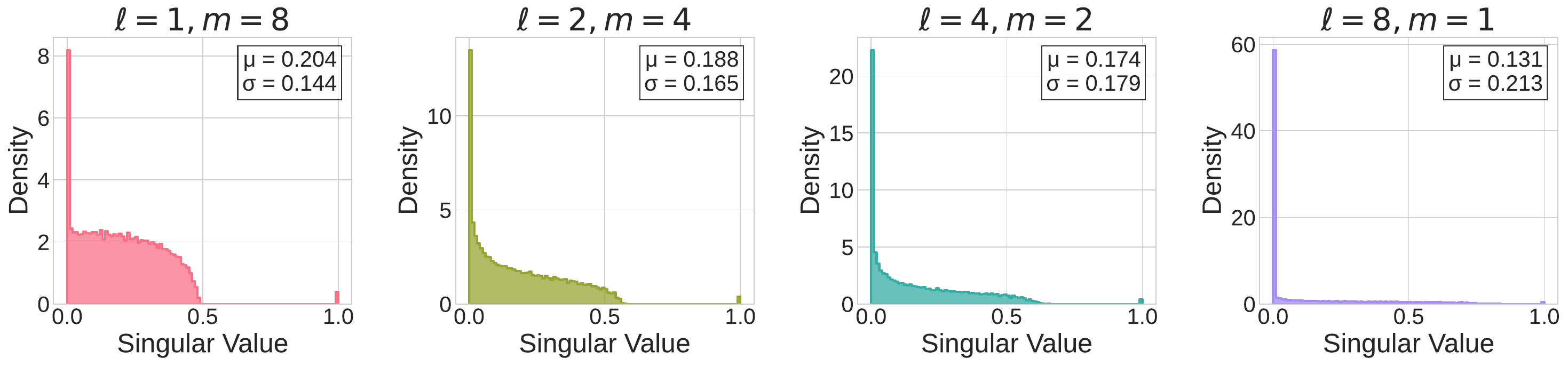}
    \caption{Empirical spectral distribution comparison with $\rep: \F_{2^m} \to \permg{16}$ and the matrix $A$. Each histogram is based on 32 trials.}
    \label{fig:eigs-d16-perm}
\end{figure}

Within the LS-MDP framework outlined in \cref{ssec:lsmdp}, the use of random permutation matrices leads to minor differences in the distribution of eigenvalues. Each permutation matrix inherently contains the fixed point $(1,1,\dots,1)$, implying it consistently possesses an eigenvalue of 1, regardless of the block matrix configuration. \cref{fig:eigs-d16-perm} illustrates the empirical eigenvalue distributions of kernels under the LS-MDP regime, highlighting variations at eigenvalue 1 when contrasted with the orthogonal case depicted in \cref{fig:partial_trans}. According to \citep{collins2014strong}, analyzing the orthogonal complement of the Perron-Frobenius eigenvector suggests asymptotic freeness, consistent with orthogonal random matrices. Nevertheless, identifying differences in eigenvalue distributions through partial transpose is equally advantageous in the context of permutations.

In summary, the partial transpose serves as a diagnostic tool for block-level dependence introduced by FRP. This observation is consistent with previous research on structured kernel analysis~\citep{huusari2021entangled} and the use of partial transposes for detecting non-separability in block matrices~\citep{peres1996separability, horodecki1997separability}. The emergence of more structured block patterns for larger $\ell$ confirms that FRP modifies the joint distribution in ways that standard random projections cannot capture.

\section{Experiments Setup for Meta RL and ICRL on POPGym}\label{sec:experiments-setup}
% Hyperparameter table
\begin{comment}
    config = {
        "LR": 5e-5,
        "NUM_ENVS": 64,
        "NUM_STEPS": 1024,
        "TOTAL_TIMESTEPS": 15e6, 
        "UPDATE_EPOCHS": 30,  // per updates
        "NUM_MINIBATCHES": 8,
        "GAMMA": 0.99,
        "GAE_LAMBDA": 1.0,
        "CLIP_EPS": 0.2,
        "ENT_COEF": 0.0,
        "VF_COEF": 1.0,
        "MAX_GRAD_NORM": 0.5,
        "ENV": AliasPrevActionV2(env),
        "ENV_PARAMS": env_params,
        "EVAL_ENV": AliasPrevActionV2(eval_env),
        "EVAL_ENV_PARAMS": eval_env_params,
        "ANNEAL_LR": False,
        "DEBUG": True,
        "S5_D_MODEL": 256,
        "S5_SSM_SIZE": 256,
        "S5_N_LAYERS": 4,
        "S5_BLOCKS": 1,
        "S5_ACTIVATION": "full_glu",
        "S5_DO_NORM": False,
        "S5_PRENORM": False,
        "S5_DO_GTRXL_NORM": False,
        "RESET_PATTERNS": (args.reset_patterns==1)
        }
    config["NUM_UPDATES"] = (
        config["TOTAL_TIMESTEPS"] // config["NUM_STEPS"] // config["NUM_ENVS"]
    ) = 228  explotation - train loop
    config["MINIBATCH_SIZE"] = (
        config["NUM_ENVS"] * config["NUM_STEPS"] // config["NUM_MINIBATCHES"]
    ) = 64*1024/8 = 8*1024
\end{comment}

\subsection{Implementation and Environment Details}\label{app:implementation}
We implemented our approach in Jax with  popjaxrl~\citep{lu2023structured} and gymnax~\citep{gymnax2022github}.
Each experiment was conducted on a computing resource under Linux OS, including 8 GPUs (NVIDIA H200 SXM5 with 150GB each) and 2 CPUs (Intel Xeon Platinum 8558 Processor). Each run used a single GPU.
We set random seeds for reproducibility and leveraged the implementation of baseline training algorithms~\citep{lu2023structured} for fairness. 

%\section{Environment Specifications}\label{app:envs}
All environments follow the POPGym configuration, except that our implementations are fully written in JAX based on \cite{lu2023structured}.

\paragraph{Stateless Cartpole}
This variant provides partial observations: the agent observes only the cart position and the pole angle, while both the linear velocity and the angular velocity are hidden. Throughout the manuscript, every figure label ``cartpole'' refers to this \emph{Stateless Cartpole} environment.

\paragraph{Higher Lower}
This is a simple card-guessing game in which the agent predicts whether the next card will be higher or lower than the current one. Observations are one-hot vectors over the $13$ card ranks. The action space has two options (predict ``higher'' or ``lower''). A correct prediction yields a positive reward whose magnitude equals the reciprocal of the number of cards in the deck; an incorrect prediction yields the same magnitude with negative sign. We use one standard deck (fifty-two cards).

\paragraph{Minesweeper}
This environment mirrors the classic Minesweeper game: the agent selects grid cells, aiming to reveal all safe cells while avoiding mines. Observations encode the count of neighboring mines as a one-hot vector. Completing the task successfully yields a positive reward equal to the reciprocal of the maximum number of steps allowed in an episode. An episode that ends in a mine hit yields a negative reward equal to one half in magnitude, further decreased by the success-reward amount. Attempting an invalid move produces a small penalty whose magnitude equals one half divided by the maximum number of steps minus two. We use a four-by-four grid with two mines.

\paragraph{Repeat First}
This memory task presents a sequence of cards. The agent must remember the suit of the very first card and choose that same suit at every subsequent step. There are four possible suits. A correct choice receives a positive reward whose magnitude equals the reciprocal of the total number of cards minus one (excluding the initial reveal), and an incorrect choice receives the same magnitude with negative sign. We use one standard deck (fifty-two cards).

\paragraph{Repeat Previous}
This is a delayed-match memory task over card suits. At each step, the agent must select the suit that appeared exactly $k$ steps earlier; the first $k$ steps carry no reward. There are four suits. For steps $k$ and beyond, a correct choice yields a positive reward whose magnitude equals the reciprocal of the total number of cards minus $k$, and an incorrect choice yields the same magnitude with negative sign. We use one deck with delay parameter $k=4$.

\subsection{Training Protocols and Duration}\label{app:training-protocol}
Our model underwent training for 228 simulation cycles, each incorporating an inner loop of 30 epochs. The optimizer employed a fixed learning rate of $5\times 10^{-5}$, with additional hyperparameters detailed in \cref{tab:hps-optimizer}. Each cycle execution required roughly 12 minutes on the specified hardware, owing to enhancements via the pure-JAX implementation by \citep{lu2023structured}.
 %and decayed it by [schedule]. %Early stopping was applied if validation performance did not improve after [P] epochs. 
%\todo{check}

\begin{table}[th]
\centering
\caption{Hyperparameter settings for Optimizer.}\label{tab:hps-optimizer}
\begin{tabular}{ccl}
\toprule
\textbf{Hyperparameter} & \textbf{Value} & \textbf{Description} \\
\midrule   
Num updates  & 228 & The number of simulation - training loops.\\
Num Epochs & 30 & The number of training epochs per updates.\\
Num Minibatches & 8 & The number of minibatches. \\
Minibatch Size              &  $8192$   & The number of 
samples per minibatch. \\
Learning Rate           &  $5 \times 10^{-5}$   & constant 
learning rate for optimizer \\
Optimizer               &  adam ($\epsilon=10^{-5}$)  & 
Optimizer for parameter updates.\\
Max Grad Norm & 0.5 & Maximum gradient norm for clipping. \\
\bottomrule
\end{tabular}
\end{table}

\subsection{Trajectory Collection Protocols and Configurations}

\cref{tab:hps-traj} lists the hyperparameters for trajectory collection. The multi environment consists of $\nenv$ sampled parallel environments. In each of these environments, a single environment step involves executing the selected action, then receiving the observation, reward, and the termination signal. When an environment concludes, a trial is added, and the environment resets to its initial state. An episode is comprised of 16 trials. Upon completing an episode, the environment resets and a new word $w$ is sampled, while the base matrices $U_1,\ldots,U_n$ defining $\lambda$ are kept fixed within the current trajectory-collection phase. The base matrices are resampled only at the beginning of the next trajectory-collection phase. In \cref{alg:fra}, $\eoe$ signifies the end flag of an episode. After Num Step parallel steps (amounting to Num Steps $\times \nenv$ in total), a trajectory collection session concludes, leading to the commencement of a model training session.

\begin{table}[th]
\centering
\caption{Hyperparameter settings for Trajectory collection from environments.}\label{tab:hps-traj}
\begin{tabular}{lcc}
\toprule
\textbf{Hyperparameter} & \textbf{Value} & \textbf{Description} \\
\midrule   
$\nenv$ (Num Envs)  & 64 & The number of parallel environments.\\
Num Steps & 1024 & The number of steps per environment per trajectory collection phase. \\
Num Trials & 16 & The number of trials per episode.%それぞれの環境の終了回数（ここに到達すればmeta envがdoneとなる）
\\
\bottomrule
\end{tabular}
\end{table}

\subsection{Random Projections Configuration}\label{ssec:rps-hps}

\cref{tab:hps-frp} shows hyperparameters for FRP.
Here, the dimension $\drp$ is chosen to be smaller than or equal to $128$, the dimension of the first hidden vector of the embedding network in \cref{ssec:arch-hps}. In the experiments in \cref{sec:pomdps}, we set $d=128$.

\begin{table}[th]
    \centering
    \caption{Hyperparameter settings  for FRP}
    \label{tab:hps-frp}
    \begin{tabular}{ccl}
        \toprule
        \textbf{Hyperparameter} & \textbf{Value} & \textbf{Description} \\
        \midrule   
        $\ell$  & $1,2,4,8$ & The word length of FRP\\
        $\scaling$ & $\sqrt{2}$ & The scaling factor of random projections\\
        $\drp$  & 128 & The dimension of FRP ($d\times d$-matrix)\\
        $\nwords$  & 256 & The number of word collections.\\
    \bottomrule
    \end{tabular}
\end{table}

\subsection{Evaluation Environment Segregation}\label{eval-environment}
It is essential to ensure that the environment used for evaluation is not included in the training environment. Random sampling of the test environment results in excessive variance, so we consider two deterministic approaches for constructing the test environment: \paragraph{Tiling} Encode the observation vector using tiling. Given an input dimension $\dobs$, replicate the input $\lfloor \din / \dobs \rfloor$ times and apply zero padding for any remaining space. Tiling is employed as discussed in \cref{sec:pomdps}. %\paragraph{Padding} Apply zero padding to the observation vector. This method is addressed in the Supplemental Experiments. 
\paragraph{Identity} Use the observation vector directly as input to the model, requiring $\din=\dobs$.

\subsection{Architecture Configuration}\label{ssec:arch-hps}

\paragraph{Partially Observable Setting.}
We apply an embedding network after the random projection. Consider $W_1$ (resp.\,$W_2$) is $128 \times d$ (resp.\,$256 \times 128$) parameter matrix initialized by iid uniform orthogonal random matrices with scaling $\sqrt{2}$. Write $\theta_e=\{ W_1, W_2\}$. 
We apply the embedding network $\EmbNN: \obs \mapsto \act\left(W_2 \act\left(W_1 \obs\right)\right)$, where $\act$ is the leaky ReLU activation function. 

The recurrent module $G_\psi$ is the composition of the embedding network and the recurrent neural network as follows: 
$\RNN(h, \obs', \eoe_t) = \RUnit(h, \EmbNN(\obs'), \eoe)$, where $\theta^r=(\theta^e, \theta^u)$. The model $\RUnit$ processes a sequence of states, updating the hidden state $h_t$ (256 dimensions) as follows: $h_t, x_t = \RNN(h_{t-1}, \obs^\prime_t, \eoe_t)$. Here, $\eoe_t$ serves as an indicator for episode termination, and $x_t$ (256 dimensions) reflects the processed state latent representation. The policy network and the value function utilize this processed state latent representation $x_t$ as input: $\pi_\theta(\cdot |\obs^\prime_t) := \PolicyNet(x_t)$ and $V_\phi(\obs^\prime_t) := \ValueNet(x_t)$, where $\theta=(\theta^r, \theta^p)$ and $\phi=(\theta^r, \theta^v)$. %In this context, $\phi$ denotes the parameters of the value function network.

\paragraph{NN configurations.}
For the module $G_\psi$ , each S5 and GRU architecture is the same as in \citep{lu2023structured}.
We summarize all key hyperparameters in \cref{tab:hps-s5}. Here full glu is the Full Gated Linear Unit, which applies both GELU activation and a sigmoid gate.  And GTRXL is the  Gated Transformer-XL.

\begin{table}[th]
    \centering
    \caption{Hyperparameter settings  for S5 Model.}
    \label{tab:hps-s5}
    \begin{tabular}{ccl}
        \toprule
        \textbf{Hyperparameter} & \textbf{Value} & \textbf{Description} \\
        \midrule   
         $\dhidden$ (D Model)  & 256 &  The feature dimension of the S5 model.\\
        SSM SIZE & 256 &  The state size of the State Space Model (SSM).\\
        N LAYERS & 4 & The number of stacked S5 layers.\\
        BLOCKS & 1 & Number of blocks to divide the SSM state into. \\
        ACTIVATION &  full glu & The activation function used in the S5 layers. \\
        DO NORM &  False &  Whether to apply layer normalization. \\
        PRENORM & False & Whether to apply normalization before or after the S5 layer.\\
        DO GTRXL NORM & False &  Whether to apply an additional normalization inspired by GTRXL.\\
         \bottomrule
    \end{tabular}
\end{table}

\subsection{PPO / GAE implementation details}\label{ssec:ppo-loss}

We use standard clipped PPO with GAE
\citep{schulman2015gae,schulman2017ppo}. The implementation follows the
recurrent PPO setup used by \citet{lu2023structured}. FRP changes only the
input projection; the optimizer, advantage estimator, and PPO objective
otherwise follow the standard modern PPO implementation
of \citet{lu2023structured}. We record the remaining switches explicitly because PPO
performance is known to depend sensitively on implementation-level choices
\citep{engstrom2020implementation,andrychowicz2021what}.

For completeness, the temporal-difference residual and GAE recursion are
\begin{align*}
\delta_t &= r_t + \gamma V_\phi(s_{t+1})(1-d_t) - V_\phi(s_t),\\
\hat A_t &= \delta_t + \gamma \lambda (1-d_t)\hat A_{t+1},
\end{align*}
where $d_t \in \{0,1\}$ is the episode-termination indicator. The actor
objective is the clipped surrogate objective of \citet{schulman2017ppo}.
Specifically, with
\[
r_t(\theta)=\frac{\pi_\theta(a_t\mid s_t)}
{\pi_{\theta,\mathrm{old}}(a_t\mid s_t)},
\]
the actor loss is
\[
L_{\mathrm{clip}}
=
-\E_t\!\left[
\min\!\left\{
r_t(\theta)\hat A_t^{\mathrm{norm}},\,
\mathrm{clip}\!\left(r_t(\theta),1-\epsilon,1+\epsilon\right)
\hat A_t^{\mathrm{norm}}
\right\}
\right],
\]
where $\hat A_t^{\mathrm{norm}}$ denotes the normalized advantage.
We normalize advantages before the policy loss and use the clipped value loss
\begin{align*}
V_\phi^{\mathrm{clip}}(s_t)
&=
V_{\phi,\mathrm{old}}(s_t)
+
\mathrm{clip}\!\left(
V_\phi(s_t)-V_{\phi,\mathrm{old}}(s_t),\,
-\epsilon,\epsilon
\right),\\
L_{\mathrm{vf}}
&=
\frac{1}{2}\,\E_t\!\left[
\max\!\left\{
(V_\phi(s_t)-R_t)^2,\,
(V_\phi^{\mathrm{clip}}(s_t)-R_t)^2
\right\}
\right],
\end{align*}
with $R_t=\hat A_t+V_{\phi,\mathrm{old}}(s_t)$. The total loss is
\[
L_{\mathrm{total}}
=
L_{\mathrm{clip}} + c_1 L_{\mathrm{vf}} - c_2 L_{\mathrm{ent}}.
\]
\cref{tab:hps-ppo} lists the PPO-specific hyperparameters and implementation
switches used in all POPGym experiments.

\begin{table}[th]
\centering
\caption{PPO-specific hyperparameters and implementation switches.}\label{tab:hps-ppo}
\begin{tabular}{lll}
\toprule
Setting & Value & Description \\
\midrule
Discount $\gamma$ & 0.99 & discount factor \\
GAE $\lambda$ & 1.0 & GAE parameter \\
Clip coefficient $\epsilon$ & 0.2 & policy/value clipping parameter \\
Entropy coefficient $c_2$ & 0.0 & entropy bonus weight \\
Value coefficient $c_1$ & 1.0 & value-loss weight \\
Advantage normalization & yes & normalize $\hat A_t$ before $L_{\mathrm{clip}}$ \\
Value loss clipping & yes & use the clipped PPO value loss \\
\bottomrule
\end{tabular}
\end{table}

\subsection{Evaluation Metrics}\label{ssec:eval}
Here, we provide a summary of the procedure for computing the ICL-Test MMER, which is the evaluation metric in \cref{sec:pomdps}. At the conclusion of each training-simulation loop, the evaluation metrics are determined as follows: The hidden state of the model is first reset. Subsequently, using the same hyperparameters as during training, as shown in \cref{tab:hps-traj}, and the evaluation environment detailed in \cref{eval-environment}, trajectories are collected. The MMER is then calculated in the same manner as for the training MMER. Once measurements corresponding to each random seed are gathered, the mean and standard deviation at each step are computed to yield the ICL-Test MMER.

\section{Implementation Details on LS-MDP}\label{sec:lsmdp-setup}

%\subsection{Enviroments}
\begin{figure}[ht]
    \centering
    \includegraphics[width=0.8\linewidth]{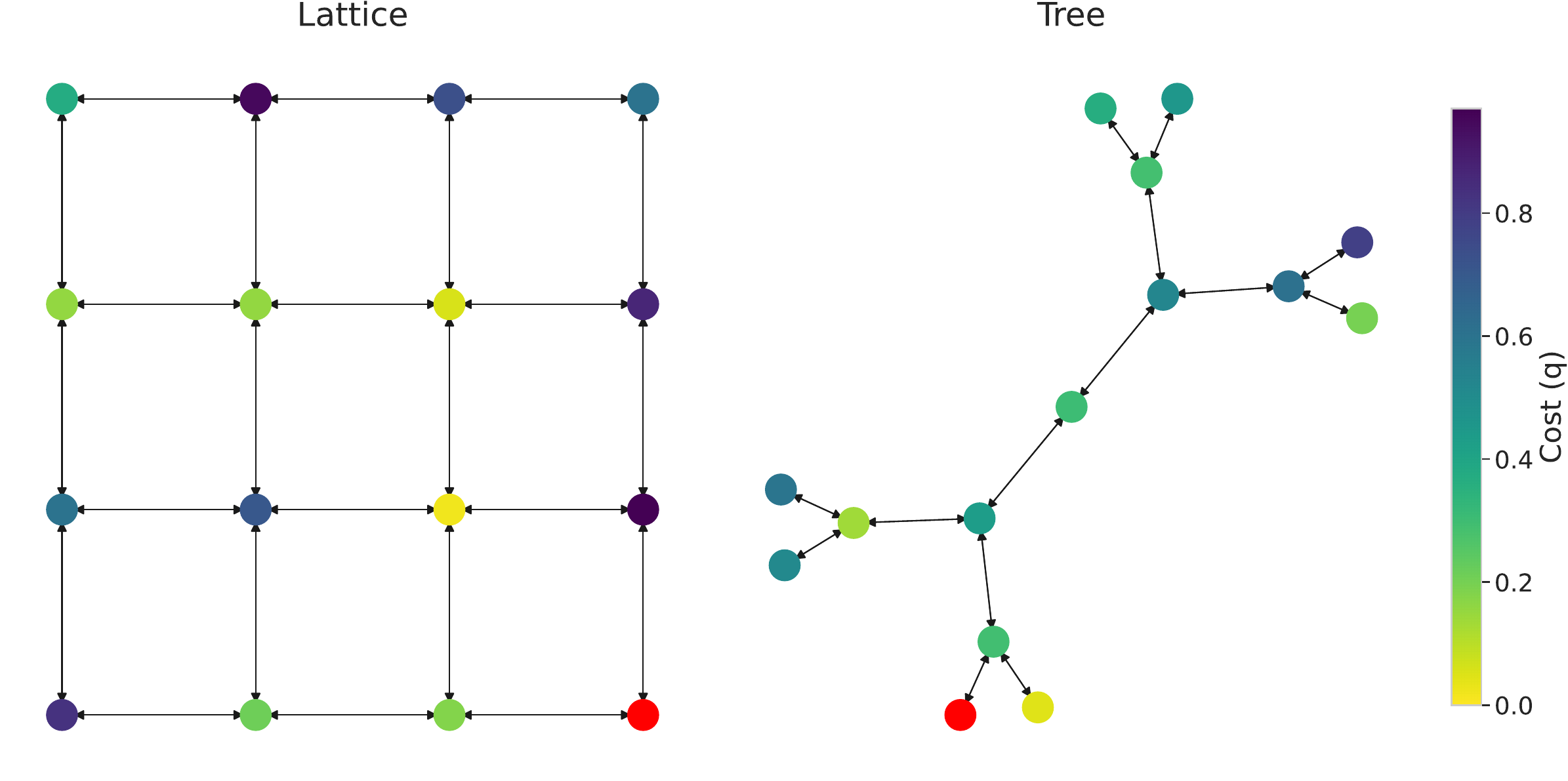}
    \caption{State spaces for LS-MDPs. The cost function is provided as an example, with vertices having zero cost highlighted in red. (Left) Lattice state space. (Right) Tree state space.}
    \label{fig:lsmdp-environments}
\end{figure}

%\paragraph{Environments}
We introduce the detailed implementation of state spaces and cost function for the LS-MDP in \cref{ssec:lsmdp}.
Define the size of the state space as $|\states|=16$ for a lattice and $|\states|=15$ for a binary tree, making them nearly equivalent. The cost distribution is applied according to $\Unif([0,1])$, with values assigned to each vertex. One extremity is set to a cost of zero artificially; for a lattice, this is one corner, and for a tree, it’s a leaf. See \cref{fig:lsmdp-environments} for examples of these environments. Transitions are only possible between connected nodes.

Within the framework of the FRP, the matrix representation $\rep: \F_n \to \permg{d}$ is determined through one sampling per experiment. We employ 10 different random seeds. As for hyperparameters, they are set as $\nwords=2^8$, $(\ell, m)=(1,8), (2,4), (4,2), (8,1)$, with $n= 2^m$, and $\drp=|\states|$.

\section{Supplemental Numerical Results}

\subsection{Impact of Word Length}\label{ssec:impact-of-length}
\begin{figure*}[t]
    \centering
    \includegraphics[width=0.245\linewidth]{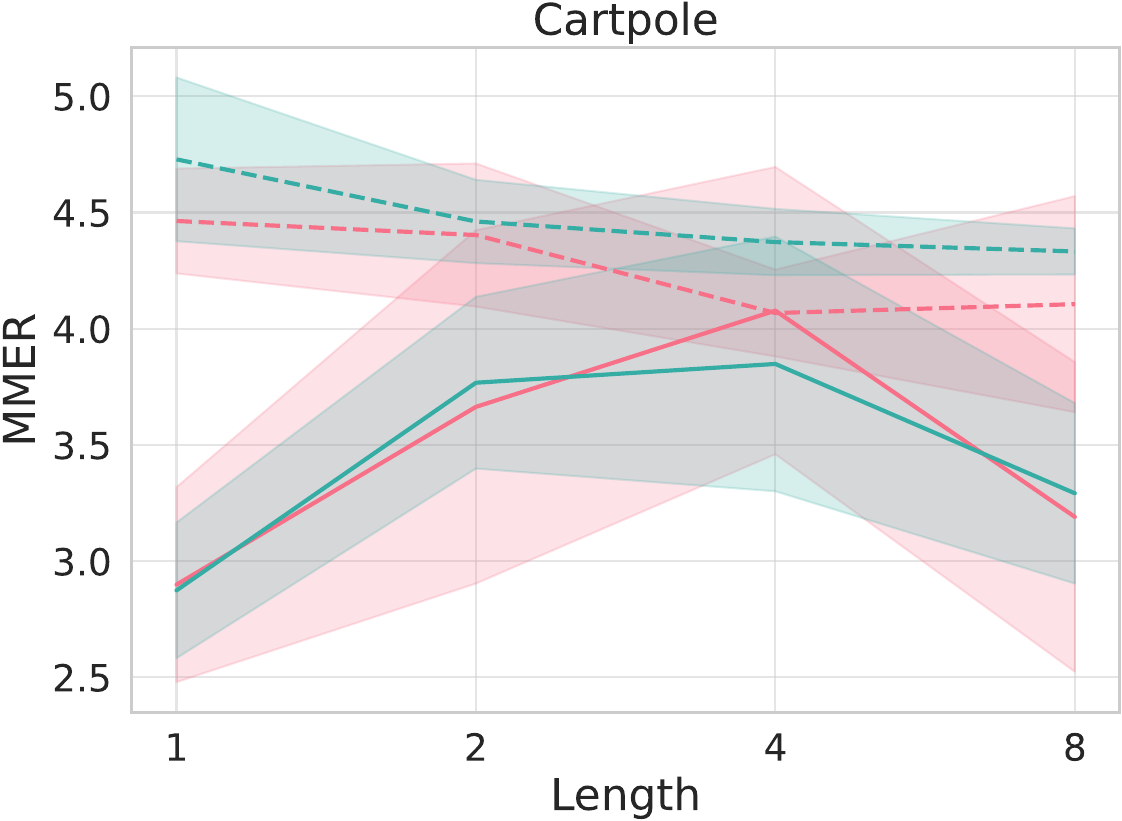}
    \includegraphics[width=0.245\linewidth]{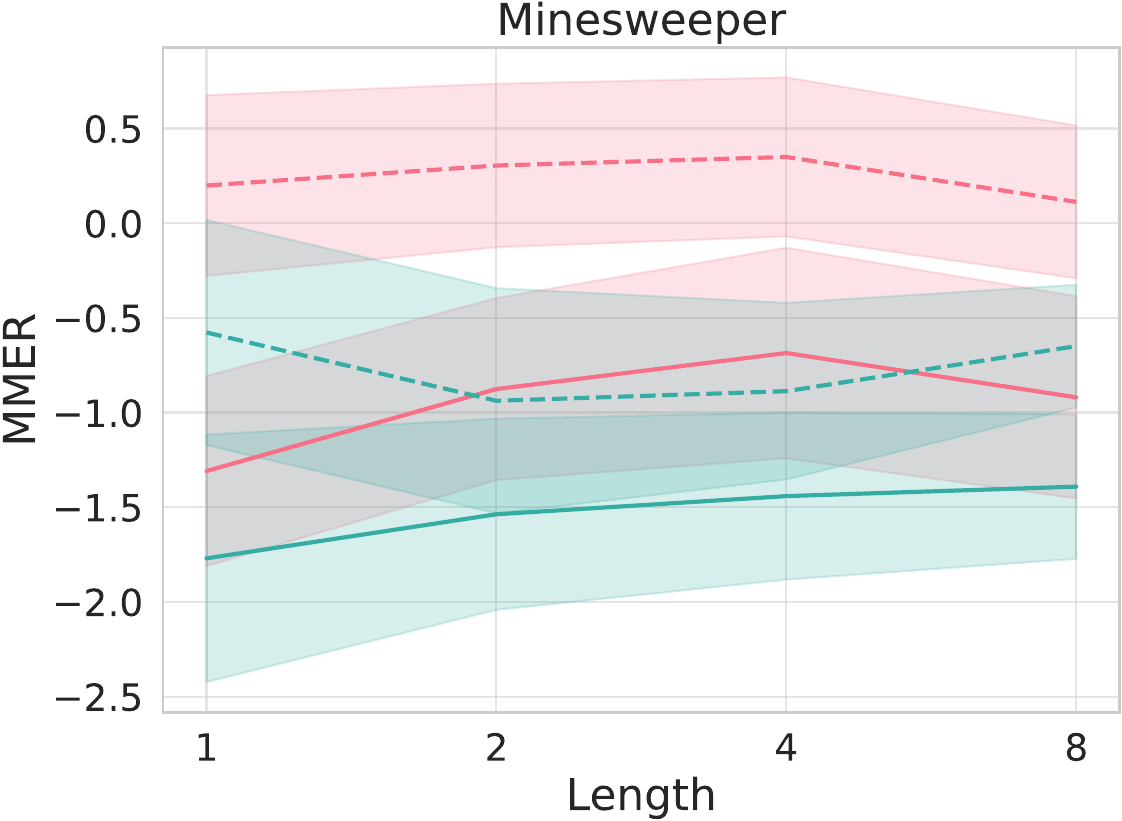}
    \includegraphics[width=0.245\linewidth]{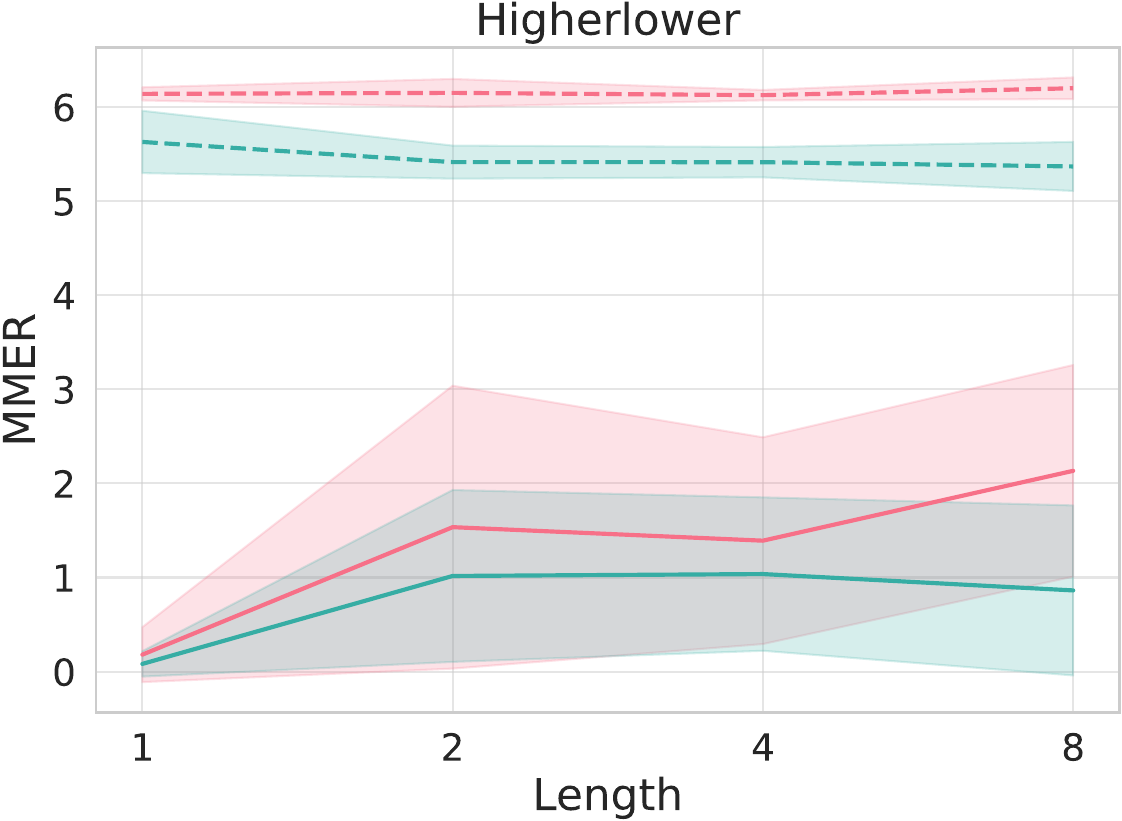}
    \includegraphics[width=0.245\linewidth]{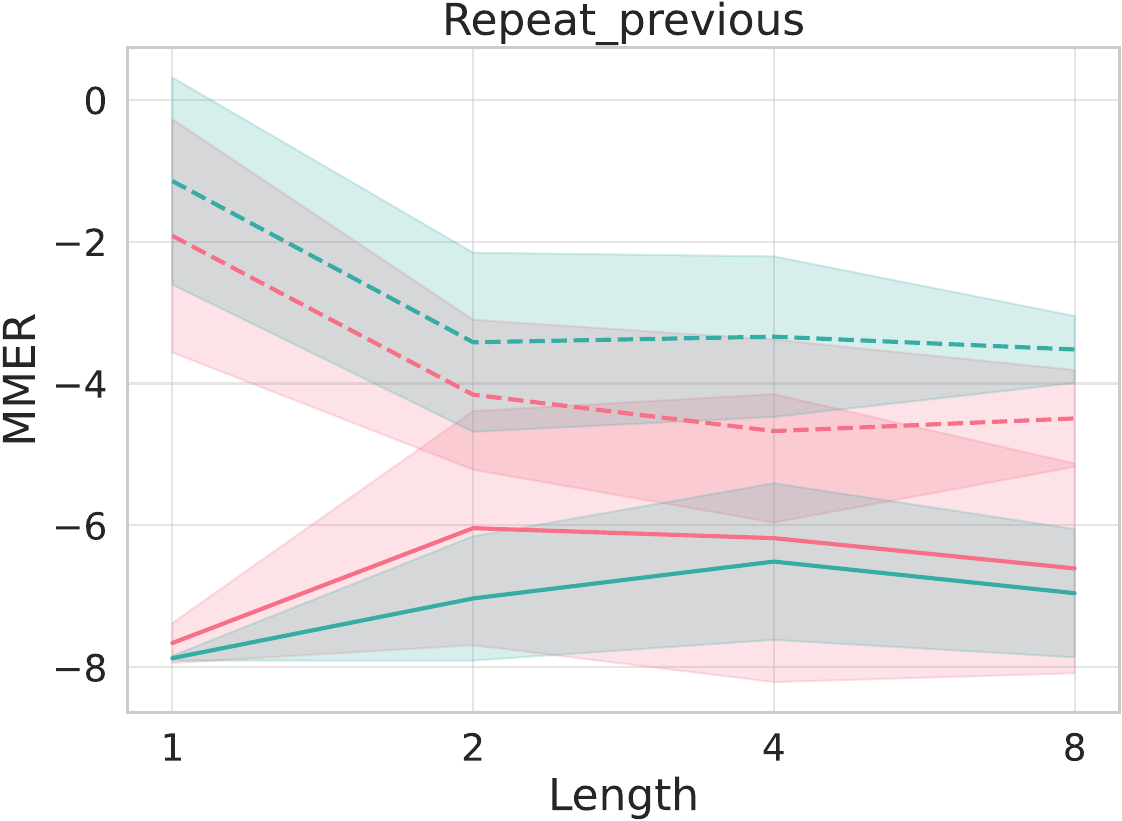}
    \includegraphics[width=0.8\linewidth]{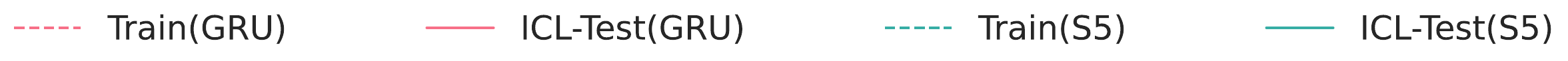}
    %(8z,128,tiling)
    %\caption{ICL-Test MMER comparison as a function of word length $\ell \in \{1,2,4,8\}$ for Stateless Cartpole, Minesweeper, Highler Lower, and Repeat Previous environments. At \( \ell=1 \), FRP collapses to the standard RP. We set $d=128$. Shaded region indicates standard error accross 10 random seeds. } 
    \caption{ICL-Test MMER vs.\ word length \(\ell\in\{1,2,4,8\}\) on Stateless Cartpole, Minesweeper, Higher Lower, and Repeat Previous. At \(\ell=1\), FRP collapses to the standard RP. \(d=128\). Mean and std over 10 seeds.}
    \label{fig:meta-envs-length}
\end{figure*}

\cref{fig:meta-envs-length} reveals a consistent pattern: in every case where \(\ell > 1\), we observe higher performance than in the standard case of \(\ell = 1\). Furthermore, all such cases show a decrease in the training MMER, indicating that the standard approach uses simpler data augmentation, which highlights the relative advantage of FRP. Increasing the word length \(\ell\) in FRP is expected to amplify the hierarchical structure in the random projections and thereby, potentially in a non-monotonic manner, improve generalization.